%% file: main.tex
\documentclass[lettersize,journal]{IEEEtran}
\usepackage{amsmath,amsfonts}
\usepackage{algorithmic}
\usepackage{algorithm}
\usepackage{array}
\usepackage[caption=false,font=normalsize,labelfont=sf,textfont=sf]{subfig}
\usepackage{textcomp}
\usepackage{stfloats}
\usepackage{url}
\usepackage{verbatim}
\usepackage{graphicx}
\usepackage{cite}
\usepackage[hidelinks]{hyperref}

\usepackage{nicematrix}
\usepackage{bm}
\usepackage{booktabs}
\usepackage{multirow}
\usepackage{bbding}
\usepackage{adjustbox}
\usepackage[normalem]{ulem}

\usepackage{CJKutf8}

\usepackage{colortbl}
\definecolor{Tabcolor}{rgb}{0.91, 0.91, 0.99}

\newcommand{\eg}{\emph{e.g.}}

\newcommand{\prompt}[1]{\textcolor{black}{\texttt{#1}}}

\begin{document}

\title{Decoupled Vision-Language System for Multimodal Understanding and Generation}

\author{Yifan Xu, Baochen Xiong, Xiaoshan Yang, Donglin Di, Yaowei Wang, Changsheng Xu
        % <-this % stops a space
% \thanks{
% Y. Xu, B. Xiong, X. Yang, and C. Xu are with MAIS, Institute of Automation, Chinese Academy of Sciences, University of Chinese Academy of Sciences, Beijing 100190, China, and also with the Peng Cheng Laboratory, Shenzhen 518066, China. Email: yifan.xu9674@gmail.com, xiongbaochen2022@ia.ac.cn, \{xiaoshan.yang, csxu\}@nlpr.ia.ac.cn. 
% }
% \thanks{Manuscript received December 25, 2025.}
% }
\IEEEcompsocitemizethanks{
\IEEEcompsocthanksitem Corresponding author: Changsheng Xu.
\IEEEcompsocthanksitem 
Yifan Xu, Baochen Xiong, Xiaoshan Yang, and Changsheng Xu are with MAIS, Institute of Automation, Chinese Academy of Sciences, University of Chinese Academy of Sciences, Beijing 100190, China, and also with the Peng Cheng Laboratory, Shenzhen 518066, China. Email: yifan.xu9674@gmail.com, xiongbaochen2022@ia.ac.cn, \{xiaoshan.yang, csxu\}@nlpr.ia.ac.cn. 
\IEEEcompsocthanksitem 
Donglin Di is with Li Auto, Beijing 101399, China. Email: donglin.ddl@gmail.com.
\IEEEcompsocthanksitem 
Yaowei Wang is with Peng Cheng Laboratory, Shenzhen 518066, China. Email: wangyw@pcl.ac.cn
}
}

% {Decoupled Vision-Language System for Multimodal Understanding and Generation}
\markboth{}%
{Decoupled Vision-Language System for Multimodal Understanding and Generation}


\maketitle

\input{sections/abstract}


\input{sections/introduction}
\input{sections/related-work}
\input{sections/preliminary}
\input{sections/method2}

\input{sections/experiments}

\input{sections/conclusion}
% \input{sections/acknowledgements}

{
\bibliographystyle{IEEEtran}
\bibliography{references}
}

\vfill

\end{document}

%% file: sections/abstract.tex
\begin{abstract}
We introduce a new architecture design for multimodal large language models (MLLMs), Libra, capable of both multimodal understanding and generation. Libra architecture contains one vision system and one language system, connected by cross-modal bridges. This design decouples self-modal modeling and cross-modal interaction, enabling each modality to learn its unique representations while maintaining effective cross-modal comprehension. The decoupling is mainly achieved in a switch attention module and a switch FFN module, which dynamically routes the computation flow for self-modal modeling and cross-modal interaction scenarios. We evaluate the effectiveness in two important settings: \textbf{Libra-1} for the understanding-only image-to-text setting, and \textbf{Libra-2} for unified image-to-text understanding and text-to-image generation. In addition to the architecture design, we discuss various improvements on tokenization, positional encoding, and supervision.
Experiments demonstrate that the dedicated Libra design enables mutual improvements on multimodal understanding and generation, achieving strong performance on both understanding and generation benchmarks. Code will be available at {\url{https://github.com/YifanXu74/Libra}}.
\end{abstract}

%% file: sections/introduction.tex
\section{Introduction} \label{sec:intro}
\IEEEPARstart{G}{enerative} modeling, particularly the auto-regressive paradigm in large language models (LLMs)~\cite{gpt3,llama3,gemini}, has emerged as a promising pathway toward general-purpose language systems over the past few years. Its effectiveness stems from an inherent capacity to reconstruct data distributions, thereby implicitly capturing the underlying world knowledge embedded in the data. Taking this a step further, in the realm of multimodal large language models (MLLMs)~\cite{libra,llava,flamingo}, an intuitive idea emerges: if a model can effectively generate text and images, it should have acquired a deeper understanding of multimodal world knowledge, while achieving stronger alignment between the two modalities. This, in turn, would lead to more robust multimodal understanding capabilities.

% % understanding-only 模型存在的必要性
% A line of MLLMs is for understanding-only image-to-text scenario,  which typically takes images and text as inputs and only supervise on the text side. 
% 图生文（理解）：信息压缩过程，将复杂的视觉像素总结为简单的语言符号，需要 非常严谨的文本叙述，较小的自由发挥空间，更难。
% 文生图（生成）：信息扩充过程，将简单的语言符号扩充为图像，有较大的自由发
% 挥空间。
% 因此，主流方法专注于解决图生文这一关键，这样训练更加高效、稳定。
% 另外一些方法致力于统一（图生文+文生图），理论上效果更佳，但是训练难度更大。
% 这两种都是重要的方向。

% Currently, there are two mainstreams in MLLMs, understanding-only models for image-to-text settings and unified models for both image-to-text and text-to-image settings. Both streams are important. 
% Image-to-text can be viewed as an information compression process: the model must distill complex visual signals into precise linguistic descriptions, where the high demand of factual response limits the room for creative deviation. By contrast, text-to-image generation is an information expansion process: simple linguistic symbols are expanded into rich visual content, allowing greater generative freedom. So, understanding-only models focus on the harder image-to-text settings, yielding more efficient and stable optimization, while unified models 理论上更scalable but  typically introduces greater training complexity and instability.

There are two representative paradigms in vision-language MLLMs: (i) understanding-only models~\cite{llava,flamingo,blip2}, which take images as conditions and generate text, namely image-to-text understanding, and (ii) unified models~\cite{unified-io,show-o,janus} for both image-to-text understanding and text-to-image generation. Image-to-text understanding is a fundamental capacity of MLLMs as it enables the model to ``see'', while text-to-image generation enables the model to learn visual world from the bottom up. Consequently, understanding-only models focus on the most fundamental image-to-text settings, often enabling more efficient and stable optimization; Unified models are, in principle, more general and scalable, but typically introducing greater training complexity and instability.

\begin{figure}[t]
    \centering
    \includegraphics[width=0.9\linewidth]{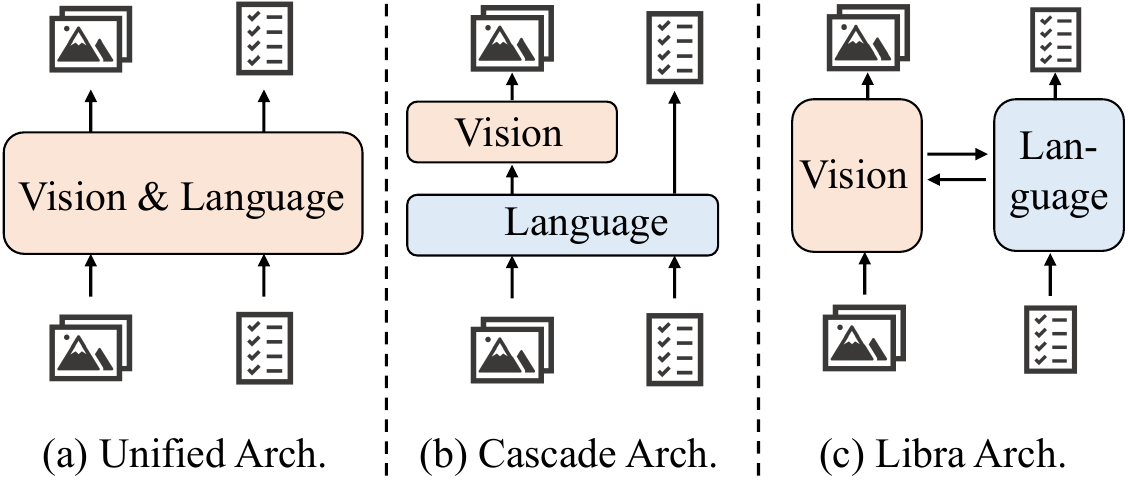}
    \caption{Architecture comparison among generative MLLMs.
    (a) A unified architecture models vision and language within a trainable shared backbone.
    (b) A cascade architecture typically uses a frozen LLM as the language backbone and adds a trainable visual backend to provide visual generation capabilities.
    (c) The Libra architecture combines the modularity of cascade architectures with the layer-wise cross-modal interaction of unified architectures.}
    \label{fig:arch_compare}
\end{figure}

Both paradigms faces a fundamental challenge in current research: the modality imbalance between vision and language.
Language data, with its highly structured symbolic representation, can compactly encode abstract concepts and relationships, while visual data organizes information in fundamentally different ways: a single image carries rich fine-grained visual details, but it has much lower semantic density than text.
As a result, existing visual datasets lag behind text corpora in the breadth of knowledge coverage, while textual datasets fall short in describing the fine-grained appearances compared to visual data.
This imbalance can easily lead to biased training, where the model favors one modality over the other, weakening performance on the less favored side~\cite{flamingo}. 
Therefore, a reasonable model design is essential to address the modality imbalance.
As shown in Fig.\,\ref{fig:arch_compare}, several types of model designs are proposed in current research:

% \begin{figure}[t]
%     \centering
%     \includegraphics[width=\linewidth]{images/radar_comp.pdf}
%     \caption{Architecture comparison among unified generative MLLMs.}
%     \label{fig:arch_comp}
% \end{figure}

% \begin{figure*}[t]
%     \centering
%     \begin{subfigure}[b]{0.48\linewidth}
%         \centering
%         \includegraphics[width=\linewidth]{images/arch_comparison.pdf}
%         \caption{Arch comparison 1.}
%         \label{fig:arch_comp1}
%     \end{subfigure}
%     \hfill
%     \begin{subfigure}[b]{0.48\linewidth}
%         \centering
%         \includegraphics[width=\linewidth]{images/image.pdf}
%         \caption{Arch comparison 2.}
%         \label{fig:arch_comp2}
%     \end{subfigure}
%     \caption{Architecture comparison among unified generative MLLMs.}
%     \label{fig:arch_comp}
% \end{figure*}

\IEEEpubidadjcol
(1) Unified architecture~\cite{llava, unified-io,chameleon,show-o,janus}, shown in Fig.\,\ref{fig:arch_compare}(a), jointly models vision and language with a shared backbone. 
Most understanding-only MLLMs~\cite{llava} are also within this architecture but are trained solely with language supervision.  This design is conceptually optimal but not well suited for the current stage of data development. 
A unified architecture, introducing minimal human prior, has the flexibility to learn arbitrary patterns for organizing vision and language, given large-scale and appropriate data. However, previous researchhas struggled with the limitations of existing vision-language datasets, which typically contain superficial knowledge and noisily aligned pairs. To overcome this, works such as Unified-IO~\cite{unified-io} and Show-O~\cite{show-o} rely heavily on large-scale pure text datasets to preserve basic language comprehension capacities. This requires vision data at a comparable scale to avoid biased training, which imposes strict constraints on data balancing.

(2) Cascade architecture~\cite{emu,dreamllm,lavit}, shown in Fig.\,\ref{fig:arch_compare}(b), decouples language and vision generation into a language backbone and a large \textit{pretrained} visual backend~\cite{ldm}. The use of the pretrained visual backend enables the pretraining of MLLMs to focus on language modeling, serving as a workaround for modality imbalance. However, a powerful visual generation backend reduces the language backbone's visual modeling to a naive role, merely serving as a prompt generator for the backend, resulting in poorly scalable multimodal understanding.

% To this end, we proposed a new type of architecture, Libra, named for its balance-scale-like design, shown in Fig.\,\ref{fig:arch_compare}(c). The core idea of Libra is to explicitly decouple both vision and language computing flows into two streams: self-modal modeling and cross-modal interaction. At the macro level, vision and language are separated into two dual branch for self-modal modeling, and connected by cross-modal bridges for layer-wise cross-modal interaction. 
% This design enables learning  from the bottom up within each modality, as in unified architectures, while mitigating modality imbalance by modularly decoupling vision and language modeling, as in cascade architectures.

% More concretely, we propose a Switch Attention to decouple the computing flows. xxx

We propose a new architecture, Libra, named for its balance-scale-like design as shown in Fig.~\ref{fig:arch_compare}(c). The core idea of Libra is to explicitly decouple both vision and language computing flows into two streams: self-modal modeling and cross-modal interaction. At the macro level, vision and language are separated into two dual branches for self-modal modeling, and connected by cross-modal bridges for layer-wise cross-modal interaction. 
This design enables learning  from the bottom up within each modality, as in unified architectures, while mitigating modality imbalance by modularly decoupling self-modal modeling and cross-modal interaction, as in cascade architectures.

% \textcolor{blue}{More concretely, the decoupling is mainly achieved in a Switch Attention, xxxx.}

% More concretely, the decoupling is mainly achieved in a proposed Switch Attention mechanism. The switch attention yield different computation patterns of the query-key dot-production for self-modal modeling and cross-modal interaction, where in self-modal modeling, it remains the standard attention, in cross-modal interaction, it map the attention key, value to a new feature space using a cross-modal bridge. Yhis strategy avoids directly aligning one modality with the feature space of another, thereby preserving the integrity of modality-specific representations while facilitating effective cross-modal interaction.

More concretely, the decoupling is primarily achieved through a proposed switch attention mechanism and a switch FFN mechanism. The switch attention induces distinct computation patterns for the query-key dot product in self-modal modeling versus cross-modal interaction. In the self-modal case, it reduces to standard attention. In the cross-modal case, it maps the attention queries, keys, and values into a new feature space via a cross-modal bridge. This strategy avoids directly aligning one modality to the feature space of another, thereby preserving modality-specific representations while enabling effective cross-modal interaction. The switch FFN also decouples multimodal understanding and generation by inducing distinct computation patterns for different scenarios.

% We evaluate the effectiveness of the Libra framework by building two variants: \textbf{Libra-1} tailored for understanding-only settings, and \textbf{Libra-2} for unified multimodal understanding and generation. 

We instantiate the Libra framework in two configurations: \textbf{Libra-1}, tailored for understanding-only tasks, and \textbf{Libra-2}, designed for unified multimodal understanding and generation.
Beyond architectural design, we introduce several new strategies for tokenization, representation, and supervision. The main contributions of this paper can be summarized as:

\begin{itemize}
    \item We introduce Libra, a new MLLM architecture whose core innovation includes a switch attention mechanism and a switch FFN mechanism, which decouple self-modal modeling from cross-modal interaction in both vision and language. This design enables the model to learn multimodal world knowledge from the bottom up while mitigating the modality imbalance that previous MLLMs suffered from. We instantiate the Libra framework in the following two configurations.
    \item \textbf{Libra-1} is designed for understanding-only tasks with two key innovations: (1) \textit{Supervision}: unified discrete autoregression over both vision and language ensures stable training; (2) \textit{Hybrid tokenization}: we integrate continuous signals with discrete IDs as vision inputs, enabling both fine-grained perception and training stability.
    \item \textbf{Libra-2} extends the framework to unified multimodal understanding and generation with three key innovations: (1) \textit{Supervision}: we integrate continuous-space masked visual generation into a unified MLLM; (2) \textit{Unified RoPE} jointly represents 2D images and 1D text while remaining natively compatible with standard RoPE; (3) \textit{Continuous-space visual tokenization}: \textbf{Libra-2 }processes only continuous visual features for both image-to-text understanding and text-to-image generation.
    \item The Libra series demonstrate strong performance on over 10 benchmarks including traditional VQA, captioning, and comprehensive MLLM benchmarks for image-to-text understanding, as well as FID metrics and general text-to-image generation benchmarks for text-to-image generation. Further analysis shows that Libra architecture exhibits lower learning redundancy than commonly used LLaVA~\cite{llava1.5} architecture.
\end{itemize}

This work extends our ICML 2024 paper~\cite{libra}, which proposed \textbf{Libra-1} that implements the framework in  understanding-only settings. In this paper, we further develop \textbf{Libra-2} that substantially advancing the Libra framework toward unified multimodal understanding and generation.

%% file: sections/related-work.tex
\section{Related Work}

In recent years, researchers have attempted to extend the LLMs to the field of multimodality. Existing model frameworks can be roughly categorized into the following two paradigms.

% \textbf{Understanding-Only MLLMs}.
% We first review early understanding-only MLLMs.
% One line of works~\cite{visual_chatgpt,gupta2023visual,shen2023hugginggpt,suris2023vipergpt,yang2023mm} utilizes LLMs as central controllers, integrating them with various functional agents, with language serving as a general interface. This plugin-style framework achieves remarkable success with very low training cost. Another line of works explores integrating pretrained vision encoders with pretrained LLMs through simple projections~\cite{llava,cogvlm,qwen-vl,otter,dreamllm,emu} or cross-attention~\cite{flamingo}.
% Several training strategies are proposed to reduce the training burden, including instruction tuning~\cite{xu2022multiinstruct,llava} and parameter-efficient tuning~\cite{lora,dettmers2023qlora,llama-adapter}.

\textbf{Unified Architecture Models}.
This category of methods performs joint modeling of vision-language modalities by sharing parameter spaces, enabling end-to-end multimodal learning. 
The advantage of this type of method lies in granting the model greater flexibility to adaptively balance visual and language modalities within a unified framework. However, this comes at the cost of complex data mixing strategies and high training expenses.
Most understanding-only and unified MLLMs are within this framework.
According to the understanding-only setting, the models are usually trained solely with language supervision, and integrate pretrained vision encoders with pretrained LLMs. The integrating strategies include simple projections~\cite{llava,cogvlm,qwen-vl,otter,dreamllm,emu}, cross-attention~\cite{flamingo}, and Q-Former prejection~\cite{blip2}.
According to the unified setting, a representative work is Unified-IO~\cite{unified-io, unified-io2}, which discretizes images into sequences of visual tokens and unifies multimodal generation tasks as sequence prediction. However, this approach requires carefully designed mixing ratios for over 20 types of multimodal task data, and it still struggles to ensure robust multimodal understanding capabilities. Subsequent studies, such as Show-O~\cite{show-o} and Janus~\cite{janus}, inject visual generation capabilities into pretrained large language models, achieving significant improvements. Nevertheless, they rely heavily on large-scale pure-text datasets to maintain language modeling performance, which not only significantly increases training costs but also necessitates fine-grained data balancing strategies. 

\textbf{Cascaded Architecture Models}.
Another category of methods adopts a staged strategy that decouples text generation and visual generation: the backbone model focuses on language modeling, while visual modeling is handled by a separate backend module. For example, Emu~\cite{emu, emu2} proposes a unified framework based on autoregressive continuous visual features and discrete language tokens. In this framework, visual generation is realized by autoregressively generating compressed one-dimensional visual features, which are then used as conditional input for a backend diffusion model to generate images. Similarly, DreamLLM~\cite{dreamllm} adopts a comparable strategy. LaViT~\cite{lavit} designs a dynamic visual encoder-decoder, which first filters and discretizes the input visual features, and then uses a diffusion model decoder to reconstruct the image. Based on this encoder-decoder, LaViT achieves joint visual and textual generation via discrete autoregression. NExT-GPT~\cite{nextgpt} further extends to multimodal generation (image, audio, video, text), employing different input encoders and designing diffusion model-based decoders for visual, audio, and video modalities to enhance generation quality.  
These methods appropriately decouple vision and language modeling at the architectural level, thus avoiding training imbalance issues. However, they usually rely on heavy backend generation modules to ensure high-quality visual outputs. This leads to a disconnect between visual modeling and language understanding: the backbone language model treats the input visual features merely as conditional signals for the backend visual generation module, and still primarily encodes visual information through language modeling approaches, making it difficult to truly internalize cross-modal knowledge.

%% file: sections/preliminary.tex
\section{Preliminaries}
\subsection{Basic Process of a Multimodal Large Language Models}
\label{sec:mllm}
MLLMs extend standard LLMs by adding visual perception to the text-only pipeline. In an LLM, raw text is tokenized into subword tokens, mapped to numeric IDs, known as tokenization. Then the IDs are embedded via a learned lookup table. A Transformer processes these embeddings in context to produce hidden representations, which a final linear classifier converts into logits over the vocabulary; applying a softmax yields next-token probabilities. MLLMs follow the same flow but augment it with an image pathway: an image encoder (\emph{e.g.}, CLIP~\cite{clip}) converts pixels into visual embeddings. The model then take a sequence that interleaves text and visual embeddings as the inputs.

Many understanding-only MLLMs~\cite{llava, qwen-vl} supervise solely on language tokens, treating image features as conditioning context. When generative modeling of images is required in a unified framework, a common strategy is to discretize input visual features with a vector-quantization process~\cite{vqgan}, producing visual token IDs so that both images and text can be modeled with the same next-token prediction objective.

% % In this work, xxx

\subsection{Lookup-free Quantization}
\label{sec:lfq}
% A quantization process of image features commonly perform matching with a set of learnable vectors defined in a vector-quantization (VQ) codebook. Specifically, a batch of images are first encoded with a vision encoder to produce the output features $f\in \mathbb{R}^{B \times H\times W \times C}$, where xxx. Then the output features are used to match with the vectors in the VQ codebook with the largest similarity:
% \begin{equation}
%     xxx
% \end{equation}
% The matched vectors are used to reconstruct the image features by training with an reconstruction loss.
% xxx

% Then each output features are us

A quantization process is used to discretize input visual features into token IDs so that both images and text can be modeled with the same next-token prediction objective, commonly performing matching with a set of learnable vectors defined in a vector-quantization (VQ) codebook. Specifically, a batch of images $I$ are first encoded with a vision encoder $\Phi_{enc}$ to produce the output features $f\in \mathbb{R}^{B \times H\times W \times C}$, where $B$ is the batch size, $H\times W$ are the spatial dimensions, and $C$ is the channel dimension. Let $Z\in\mathbb{R}^{B\times N\times C}$ with $N=HW$ denote the spatially flattened features, and let $\mathbf{E}=\{e_k\}_{k=1}^K\in\mathbb{R}^{K\times C}$ be a learnable VQ codebook.
Then the output features are used to match with the vectors in the VQ codebook with the largest similarity:
\begin{equation}
\begin{aligned}
q_{b,n} &= \arg\max_{q\in\mathbf{E}} \frac{\langle z_{b,n}, q\rangle}{\|z_{b,n}\|_2\,\|q\|_2},
\end{aligned}
\end{equation}
where $z_{b,n}\in\mathbb{R}^C$ is the feature at location $n$ of sample $b$, and $q_{b,n}$ is its quantized counterpart. The matched vectors are used to reconstruct the images by training with a reconstruction loss, \emph{e.g.},
$\mathcal{L}_{\mathrm{rec}}=\|I-\Phi_{dec}(q)\|_2^2$, where $\Phi_{dec}$ is a decoder to decode the quantized counterpart $q$ into the pixel space.

Compared with previous quantization methods~\cite{vqgan, residualvq, fsq}, Lookup-free Quantization (LFQ)~\cite{lfq} reduces the embedding dimension of the VQ codebook~\cite{vqvae} to zero. Specifically, the codebook $\mathbf{E} \in \mathbb{R}^{K \times C}$, similar to the one in VQGAN~\cite{vqgan}, is replaced with an integer set $\mathbb{E}$ where $|\mathbb{E}|=K$, where $K$ is the vision vocabulary size. This approach eliminates the need for embedding lookup entirely. Unlike previous quantization methods~\cite{vqgan,vqvae} that require codebook embeddings to mimic input features for image reconstruction, LFQ does not require such emulation as it has no codebook embeddings.

% In this work, xxx

\begin{figure*}
    \centering
    \includegraphics[width=\textwidth]{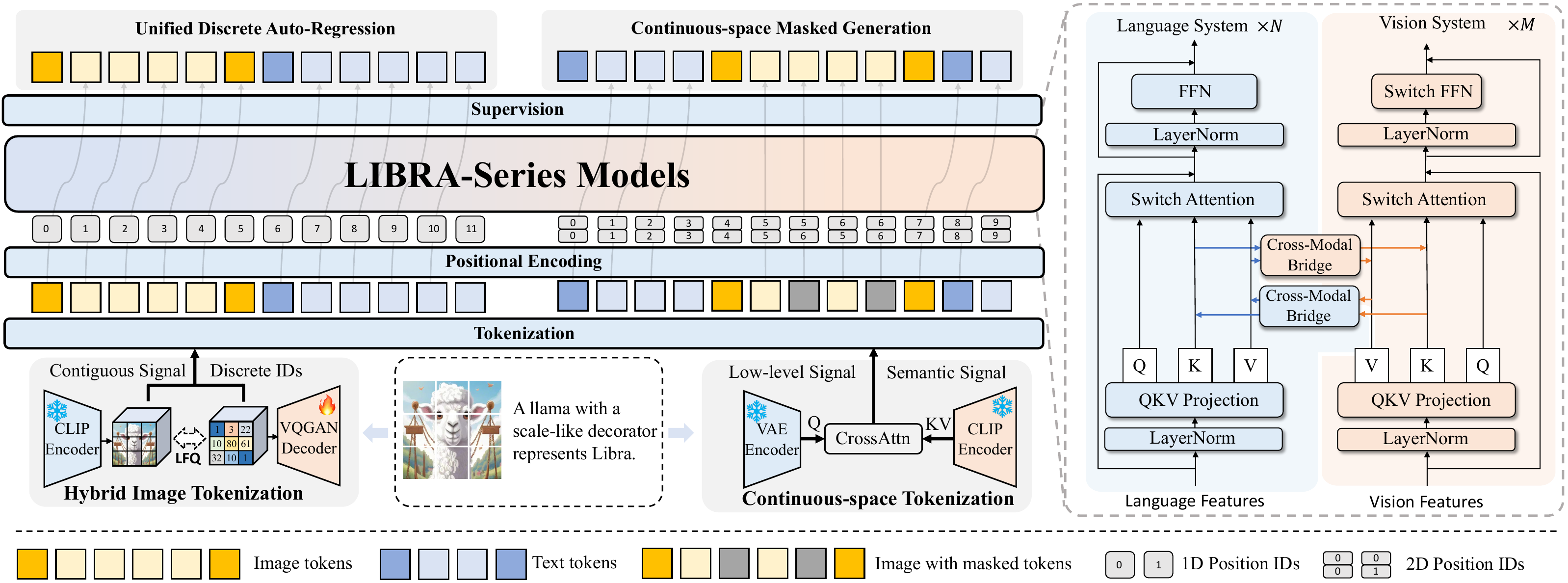}
    \caption{An overview of the Libra architecture. The core design is the switch attention and the switch FFN that decoupling self-modal modeling and cross-modal interaction. We build two variants, \textbf{Libra-1} and \textbf{Libra-2}, to discuss several improvements in tokenization, positional encoding, and supervision.}
    \label{fig:overview}
\end{figure*}

\subsection{Rotary Position Embedding}
\label{sec:rope}
% Rotary Position Encoding (RoPE)~\cite{rope} is a technique that encoding the positional information of tokens within a Transformer attention, which is widely used in current LLM frameworks. 
% For the $i$-th token in a sequence, RoPE first define a matrix $R_{\Theta, i}^{d}$ to represent its position:
% \begin{equation*}
% \begin{adjustbox}{max width=\linewidth}
% $
% \setlength{\arraycolsep}{2pt}
% \renewcommand{\arraystretch}{1.2}
% R_{\Theta, i}^{d}=
% \begin{bmatrix}
% \cos(i\theta_0) & -\sin(i\theta_0) & & & & & \\
% \sin(i\theta_0) &  \cos(i\theta_0) & & & & & \\
% & & \cos(i\theta_1) & -\sin(i\theta_1) & & & \\
% & & \sin(i\theta_1) &  \cos(i\theta_1) & & & \\
% & & & & \ddots & & \\
% & & & && \cos(i\theta_{d/2}) & -\sin(i\theta_{d/2}) \\
% & & & && \sin(i\theta_{d/2}) &  \cos(i\theta_{d/2}) \\
% \end{bmatrix}
% $
% \end{adjustbox}
% \label{eqn:unified_rope}
% \end{equation*}
% where $\Theta=\left\{\theta_i=10000^{-2(i-1) / d}, i \in[1,2, \ldots, d / 2]\right\}$ is the set of predefined frequencies, and $d$ is the channel dimension. The RoPE matrix must satisfy relative equivalence:
% \begin{equation}
%     R_{\Theta, i}^{d} R_{\Theta, j}^{d} = R_{\Theta, i-j}^{d}
% \end{equation}

% Based on this, the attention interaction of the $i$-th token  and the $j$-th token can be:
% \begin{equation}
%     (Q_{i} R_{\Theta, i}^{d}) (K_j R_{\Theta, j}^{d})^{T} = Q_{i} R_{\Theta, i-j}^{d} K_j^{T}
% \end{equation}
% where xxx. It can be xxxx.

% RoPE introduction
Rotary Position Embedding (RoPE)~\cite{rope} encodes token positions by rotating
query and key vectors within each Transformer head.
Let $d$ be the (even) channel dimension of a head. For the token at position $i$,
define the block-diagonal rotation matrix $R_{\Theta,i}^{(d)} \in \mathbb{R}^{d \times d}$ as
\begin{equation*}
\begin{adjustbox}{max width=\linewidth}
$
\setlength{\arraycolsep}{2pt}
\renewcommand{\arraystretch}{1.2}
R_{\Theta, i}^{d}=
\begin{bmatrix}
\cos(i\theta_0) & -\sin(i\theta_0) & & & & & \\
\sin(i\theta_0) &  \cos(i\theta_0) & & & & & \\
& & \cos(i\theta_1) & -\sin(i\theta_1) & & & \\
& & \sin(i\theta_1) &  \cos(i\theta_1) & & & \\
& & & & \ddots & & \\
& & & && \cos(i\theta_{d/2}) & -\sin(i\theta_{d/2}) \\
& & & && \sin(i\theta_{d/2}) &  \cos(i\theta_{d/2}) \\
\end{bmatrix}
$
\end{adjustbox}
\label{eqn:rope}
\end{equation*}
where the per-pair angular frequencies are collected in
\begin{equation*}
\Theta=\bigl\{\theta_m = b^{-2m/d} \;:\; m=0,1,\ldots,\tfrac{d}{2}-1\bigr\}, \quad b=10000.
\end{equation*}
Each $2\times 2$ block rotates the coordinates of a feature pair $(2m,2m+1)$ by angle $i\theta_m$. These rotations satisfy the relative-shift property:
\begin{equation}
R_{\Theta,i}^{(d)}\bigl(R_{\Theta,j}^{(d)}\bigr)^{\!\top} = R_{\Theta,i-j}^{(d)}.
\label{eq:rope-rel}
\end{equation}

RoPE is applied to queries and keys before their dot-product interaction. Given
$Q_i, K_j \in \mathbb{R}^{1\times d}$ for the query/key row vectors at positions $i$ and $j$,
\begin{equation}
\bigl(Q_i R_{\Theta,i}^{(d)}\bigr) \bigl(K_j R_{\Theta,j}^{(d)}\bigr)^{\!\top}
\;=\;
Q_i \, R_{\Theta,i-j}^{(d)} \, K_j^{\top},
\label{eq:rope-attn}
\end{equation}
so the attention score depends only on the relative position $i-j$ rather than
absolute positions. In conclusion, RoPE provide a simple and parameter-free way to endow attention with
relative positional inductive bias.

% In this work, xxx

\subsection{Masked Image Generation in Continuous Space}
\label{sec:mar}
% Masked generative modeling~\cite{maskgit, mage} is a paradigm for image generation that learns to predict randomly masked tokens from their unmasked context. Let $X=\{x_i\}_{i=1}^{N}$ denote the sequence of spatial tokens for an image, where each token $x_i \in \mathbb{R}^D$ may be continuous or discrete (\emph{e.g.}, a codebook index). Masked generative modeling introduces a learnable special token $x_\ast$ used to indicate missing content, as:
% \begin{equation}
%     p(x_{1}, x_{2}, \cdots, x_{N}) = \prod_{i=1}^{N} p(x_{i} | x_*, x_2, \cdots, x_*, \cdots x_{N})
% \end{equation}

% MAR~\cite{mar} first introduced masked image generation in continuous space, where each masked token is predicted by a light-weight diffusion head. The training process is:
% 1) Patchify the images into input tokens of a Transformer with a VAE~\cite{vae}, and get $X=\{x_i\}_{i=1}^{N}$.
% 2) Randomly mask the input tokens with $x_*$, as $X'=\{x_*, x_2, \cdots, x_*, \cdots x_{N}\}$.
% 3) Feed the partially masked $X'$ into the main Transformer model $\Phi$, and get output features $\{f_*, f_2, \cdots, f_*, \cdots f_{N}\}$.
% 4) Predict the corresponding original features with the masked features using $L_{diff}(\Phi_{dec}(f_*), x_k)$, where $L_{diff}$ is the diffusion loss, can be trivially treated as a reconstruction loss, $\Phi_{dec}$ is a light-weight diffusion head for predicting each individual tokens and do not need to see the whole image.

% The inference process is xxx.

% \paragraph{Masked generative modeling in continuous space.}
Masked generative modeling~\cite{maskgit, mage} learns an image prior by predicting randomly masked tokens from their visible context. Let $X=\{x_i\}_{i=1}^{N}$ denote the sequence of spatial tokens for an image, where each token $x_i \in \mathbb{R}^D$ may be continuous (e.g., from a latent encoder) or discrete (e.g., a codebook index). A learnable special token $x_\ast$ indicates missing content. For a mask set $M \subseteq [N]$, define the masked input
\[
X^{(M)}_i \;=\;
\begin{cases}
x_\ast, & i \in M \\
x_i, & i \notin M \;.
\end{cases}
\]
Masked modeling maximizes a pseudo-likelihood objective that predicts each masked token from its unmasked context:
\begin{equation}
\label{eq:mlm}
\mathcal{L}_{\text{MIM}}(\theta)
\;=\;
\mathbb{E}_{X \sim p_{\text{data}}}\;
\mathbb{E}_{M \sim \pi}\;
\sum_{i \in M}
% \mathbb{E}_{i \in M}\;
\log p_\theta\!\bigl(x_i \mid X^{(M)}\bigr),
\end{equation}
where $\pi$ is a masking distribution. This objective reduces to cross-entropy for discrete $x_i$ and to reconstruction-style losses for continuous $x_i$.

MAR~\cite{mar} introduces masked image generation in continuous space by attaching a lightweight diffusion head to each position. Training proceeds as follows:
\begin{enumerate}
\item Encode an image with a VAE~\cite{vae} to obtain tokens $X=\{x_i\}_{i=1}^{N}$ with $x_i \in \mathbb{R}^D$.
\item Sample a random mask set $M \subseteq [N]$ and construct $X^{(M)}$ by replacing $\{x_i : i \in M\}$ with $x_\ast$.
\item Feed $X^{(M)}$ to the Transformer backbone $\Phi$ to produce per-position features $F=\Phi\!\bigl(X^{(M)}\bigr)$. Let $f^{(i)} \in F$ denote the feature at position $i$; for masked positions $i \in M$, write $f_\ast^{(i)}$.
\item For each $i \in M$, apply the per-token diffusion head $\Phi_{\text{dec}}$ to $f_\ast^{(i)}$ and minimize the diffusion loss
$\mathcal{L}_{\text{diff}}\!\bigl(\Phi_{\text{dec}}(f_\ast^{(i)}),\, x_i\bigr)$,
which can be viewed as a reconstruction-style objective.
\end{enumerate}

At inference time, MAR starts from a fully masked sequence and generates iteratively. In each iteration, it predicts all tokens, then randomly accepts a fixed proportion of the newly predicted tokens, re-masks the remainder, and repeats until all positions are accepted.

% In this work, 

%% file: sections/method2.tex
\section{Decoupled Vision-Language System}
\label{sec:arch}
% \subsection{Overview}
Libra framework is built on three basic principles:
\begin{enumerate}
    \item \textit{Modularity}: The vision and language systems should be partially decoupled, thereby mitigating biased learning that results from modality entanglement.
    \item \textit{Interactivity}: A well-designed interaction mechanism is essential to connect the two systems and achieve effective cross-modal interaction.
    \item \textit{Integrity}:  Each modality should be modeled and learned end-to-end from the bottom up to ensure deep and complete understanding.
\end{enumerate}

Fig.\,\ref{fig:overview} shows an overview of the Libra framework. This framework comprises a language system and a vision system (\textit{Modularity}), connected via cross-modal bridges for interaction (\textit{Interactivity}). Each system serves as a modality-specific generator, learning in-depth world knowledge in a bottom-up manner (\textit{Integrity}).

The methodology part is organized as follows. Sec.~\ref{sec:arch} (this section) presents the overall pipeline and the core architectural design of the Libra framework. Since understanding-only and unified models require tailored tokenization and supervision strategies, we instantiate distinct configurations to build \textbf{Libra-1} for understanding-only MLLMs in Sec.\ref{sec:libra1} and \textbf{Libra-2} for unified MLLMs in Sec.\ref{sec:libra2}.

\subsection{Overall Pipeline}
Libra’s pipeline comprises several components: tokenization, forward pass, positional encoding, and supervision. Here we present a meta-framework to illustrate the overall functionality. Concrete instantiations are provided in Sec.~\ref{sec:libra1} for understanding-only settings and Sec.~\ref{sec:libra2} for unified settings.

\textbf{Tokenization and Forward Pass}. The tokenization and forward pass of Libra follow a similar process to that of a basic MLLM introduced in Sec.~\ref{sec:mllm}. For coherence, we reiterate the process here.
Libra extends standard LLMs by adding visual perception to the text-only pipeline. In an LLM, raw text is tokenized into subword tokens, mapped to numeric IDs, known as tokenization. Then the IDs are embedded via a learned lookup table. A Transformer~\cite{transformer} processes these embeddings in context to produce hidden representations, which a final linear classifier converts into logits over the vocabulary; applying a softmax yields next-token probabilities. Libra follows the same flow but augment it with an image pathway: an image tokenization process converts pixels into visual embeddings. The model then take a sequence that interleaves text and visual embeddings as the inputs.

\textbf{Positional Embedding}.
Libra adopts rotary positional embeddings (RoPE)~\cite{rope}, introduced in Sec.~\ref{sec:rope}. To unify positional representations for 2D images and 1D text within a single model, we propose a unified RoPE (UniRoPE) method, described in Sec.~\ref{sec:libra2}.

\textbf{Supervision}.
Libra conducts supervision on both the image and language sides for both the understanding-only setting and the unified-generation setting. On the language side, Libra uses vanilla next-token prediction. On the vision side, we instantiate different modeling strategies: discrete auto-regression in Sec.~\ref{sec:libra1} for the understanding-only setting, and continuous-space masked generation in Sec.~\ref{sec:libra2} for unified generation.

\subsection{Core Architecture}
\label{sec:core_arch}
\textbf{Switch Attention}.
\label{sec:switch_attn}
% why use switch attention
The interaction between the two systems in Libra primarily occurs within the attention mechanism, where each token dynamically switches its computational pathway between self-modal modeling and cross-modal interaction. We refer to this as switch attention. This mechanism helps prevent self-modal generation from interfering with cross-modal comprehension during training.

Given a multimodal input consisting of image and text features with channel dimension $D$, denoted as $X=[X_{I}, X_{T}]$, the switch attention is computed as:

\begin{equation}
\begin{aligned}
\operatorname{Attn}(X) &= \operatorname{softmax}\left(\frac{F_b(Q, K)}{\sqrt{D}}\right) F_b'(V), \\[1.5ex]
Q &= \left[Q_I, Q_T\right],\\
K &= \left[K_I, K_T\right],\\
V &= \left[V_I, V_T\right],
% \text{where} \quad Q_{*} = X_{*} W_{*}^Q, \quad K_{*} = X_{*} W_{*}^K, \quad V_{*} = X_{*} W_{*}^V, \quad * \in \{I, T\}.
\end{aligned}
\label{eqn:libra2_attn_expert}
\end{equation}
where $Q$, $K$, $V$ are projections of $X$.
% where the attention mask is omitted and explained in Sec.\,\ref{}. 
The cross-modal bridges $F_b$ and $F_{b}^{\prime}$ are the core design, as:
\begin{equation}
\begin{aligned}
    F_b(Q, K) &= \left[\begin{array}{ll}
    Q_I K_I^{\top} & Q_I^{\prime} K_T^{\prime \top} \\
    Q_T^{\prime} K_I^{\prime\top} & Q_T K_T^{\top}
    \end{array}\right], \\
    F_{b}^{\prime}(V) &= \left[\begin{array}{ll}
    V_I & V_I^{\prime} \\
    V_T & V_T^{\prime}
    \end{array}\right],
    % Q_{*}^{\prime} = Q_{*} + Q_{*} W_{*}^{Q^{\prime}}, \quad K_{*}^{\prime} = K_{*} W_{*}^{K^{\prime}}, \quad V_{*}^{\prime} = V_{*} W_{*}^{V^{\prime}}, \quad * \in \{I, T\}.
\end{aligned}
\label{eqn:cross_bridge_v2}
\end{equation}
where 
\begin{equation}
    \begin{aligned}
        Q_{*}^{\prime} &= Q_{*} + Q_{*} W_{*}^{Q^{\prime}},\\
        K_{*}^{\prime} &= K_{*} W_{*}^{K^{\prime}},\\
        V_{*}^{\prime} &= V_{*} W_{*}^{V^{\prime}},\\
        * &\in \{I, T\}.
    \end{aligned}
\end{equation}
Here, the projection matrices $W_{*}^{Q'}$, $W_{*}^{K'}$, $W_{*}^{V'}$ are the cross-modal bridges. $W_{*}^{K'}$ and $W_{*}^{V'}$ are standard matrix in a linear layer, while $W_{*}^{Q'}$ is a low-rank matrix of rank $8$, serving as a low-rank adaptation~\cite{lora} applied to $Q$ during cross-modal interactions. This enables lightweight finetuning of the original modality-specific query patterns, rather than learning entirely new ones, preserving modality-specific feature structures while enhancing cross-modal alignment.
% As a result, the approach preserves modality-specific feature structures while enhancing cross-modal alignment.

An intuitive explanation of the switch attention mechanism is as follows: For self-modal modeling (\emph{e.g.}, $Q_I K_I^{\top}$), Libra retains the standard self-attention operation. In contrast, for cross-modal interaction (\emph{e.g.}, $Q_I' K_T'^{\top}$), it redirects the computation pathways of $Q$, $K$, and $V$, projecting them into a new latent space via cross-modal bridges before performing attention alignment. This strategy avoids directly aligning one modality with the feature space of another, thereby preserving the integrity of modality-specific representations while facilitating effective cross-modal interaction.

\begin{figure}[t]
    \centering
    \includegraphics[width=\linewidth]{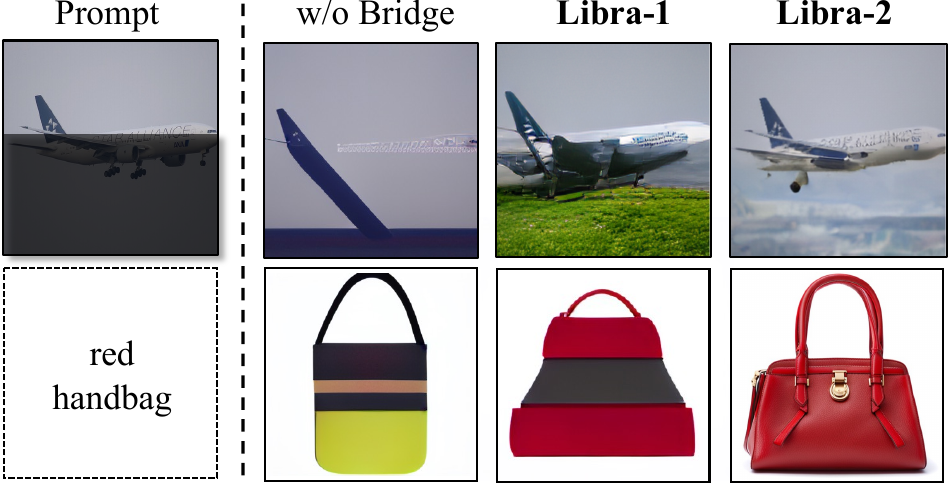}
    \caption{Assessing the cross‑modal bridge via image inpainting and text‑to‑image generation. We finetune \textbf{Libra-1} and its variant without cross-modal bridges with additional 10M image-text pairs from the
pretraining data by disabling the contiguous visual signal in Sec.~\ref{sec:supervision}. We also present the results of \textbf{Libra-2} for reference.}
    \label{fig:cross_modal_bridge}
\end{figure}

\textbf{Cross-modal Bridge}.
The cross-modal bridge in Eqn.~\eqref{eqn:cross_bridge_v2} is the core component of the switch attention mechanism. Here we further elaborate on its role in mitigating modality imbalance. Rather than forcing one modality to align directly to the feature space of another, the bridge serves as an intermediate adapter that preserves modality-specific characteristics while enabling controlled information exchange. As shown in Fig.~\ref{fig:cross_modal_bridge}, removing the bridge yields an entangled and underperforming vision system that (1) learns repetitive patterns in image completion and (2) poorly follows language instructions in text-to-image generation. In contrast, the bridge implements a principled cross-modal interaction strategy that substantially strengthens visual learning within the MLLM.

\textbf{Switch FFN}.
Multimodal understanding and generation are distinct processes. While understanding concerns the comprehension of overall meaning, generation emphasizes the production of detailed content. These two processes should be treated in different manners. 
Inspired by switch attention, we further propose switch FFN that dynamically routes computation in the visual branch to decouple multimodal understanding from generation. The realization of switch FFN is simple. Given image features $X_I = \{X_I^{\text{und}}, X_I^{\text{gen}}\}$, where $X_I^{\text{und}}$ and $X_I^{\text{gen}}$ are the feature subsets dedicated to understanding and generation, switch FFN applies separate FFNs to these subsets and concatenates the results:
\begin{equation}
    SwichFFN(X_{I}) = concat(FFN_{u}(X_{I}^{und}), FFN_{g}(X_{I}^{gen})).
\end{equation}
In words, if the image features are associated with understanding, it is routed to an understanding expert $FFN_{u}$, and vice versa. Together with switch attention, the simple yet effective switch FFN further strengthens the decoupling, leading to improved performance in both understanding and generation. 
We use the switch FFN only in the vision system. This prevents the pretrained LLM’s weights in the language system from being significantly disturbed, as no additional parameters are added to its forward pass.
The final FFN for image and text features $X=\{X_I, X_T\}$ is computed as:
\begin{equation}
    FFN(X) = concat(SwichFFN(X_{I}), FFN(X_T)).
    \label{eqn:switch_ffn}
\end{equation}

% MLLMs extend standard LLMs by adding visual perception to the text-only pipeline. In an LLM, raw text is tokenized into subword tokens, mapped to numeric IDs, known as tokenization. Then the IDs are embedded via a learned lookup table. A Transformer processes these embeddings in context to produce hidden representations, which a final linear classifier converts into logits over the vocabulary; applying a softmax yields next-token probabilities. MLLMs follow the same flow but augment it with an image pathway: an image encoder (\emph{e.g.}, CLIP~\cite{clip}) converts pixels into visual embeddings. The model then take a sequence that interleaves text and visual embeddings as the inputs.

% Many understanding-only MLLMs~\cite{llava, qwen-vl} supervise solely on language tokens, treating image features as conditioning context. When generative modeling of images is required in a unified framework, a common strategy is to discretize input visual features with a vector-quantization process~\cite{vqgan}, producing visual token IDs so that both images and text can be modeled with the same next-token prediction objective.

\section{Libra-1: Towards Complete Multimodal Understanding}
\label{sec:libra1}
\subsection{Model Overview}

Multimodal understanding, typically in image-to-text settings, is the fundamental ability of MLLMs.
Here we built \textbf{Libra-1} to validate the effectiveness of the Libra framework under understanding-only image-to-text settings.

\textbf{Libra-1} follows the core architecture described in Sec.~\ref{sec:core_arch}. Although the model targets understanding only, we supervise both the image and text sides to meet Libra’s integrity principle that each modality should be modeled and learned end-to-end from the bottom up to ensure deep and complete understanding. Since \textbf{Libra-1} does not generate images, we replace the switch FFN in Eqn.~\eqref{eqn:switch_ffn} with a standard FFN, as:
\begin{equation}
\operatorname{FFN}(X)
=
\operatorname{concat}\!\left(
  \operatorname{FFN}_I\!\left(X_I\right),\,
  \operatorname{FFN}_T\!\left(X_T\right)
\right).
\label{eq:ffn_expert}
\end{equation}

The following subsections detail the tokenization and supervision of \textbf{Libra-1}.

\subsection{Hybrid Tokenization}
\label{sec:tokenization}
\textbf{Libra-1} unifies both vision and language modeling into a \textit{discrete} next-token-prediction paradigm. Given an input sequence with both image and corresponding language parts, we separately tokenize the image and the language parts into discrete tokens through a VQGAN~\cite{vqgan} and a SentencePiece~\cite{sentencepiece} tokenizer. We respectively prefix and suffix the image sequence with one $\langle \textrm{BOI} \rangle$ (beginning of image) token and one $\langle \textrm{EOI} \rangle$ (end of image) token. 
We use a newline token ``\verb|\n|'' to separate images and texts.
All token embeddings except the separation newline token ``\verb|\n|'' are updated through a cross-entropy classification loss.
% Meanwhile, integrating images into completely discrete tokens results in severe information loss, as verified in Sec.\,\ref{sec:discuss}. Therefore, we propose a hybrid image tokenization process (Sec.\,\ref{}) to enable both stable discrete sequential modeling and contiguous visual comprehension.
% 
% 
% \subsubsection{Hybrid Tokenization}
% \label{sec:tokenization}

This unified discrete next-token-prediction paradigm enables stable training convergence but raising two obstacles for effective vision-language comprehension: 1) The discretization process of VQGAN can cause severe visual information loss, leading to low perception on visual details. 2) Naive discrete sequential modeling hardly benefits from the pretrained knowledge of the vision encoder, since the model receives newly-constructed embeddings based on the input IDs instead of the features of the vision encoder. 
To this end, we propose a hybrid image tokenization process from two aspects: contiguous visual signals and pretrained visual knowledge.

\textbf{Contiguous Visual Signal vs. Discrete Modeling.}
We leverage a hybrid tokenization strategy with a combination of contiguous visual signals from the vision encoder and discrete modeling using tokenized IDs. Formally:
\begin{equation}
    \begin{aligned}
        f_c &= \Phi(I), \\
        \text{ID} &= \operatorname{quantize}(f_c), \\
        f_d &= E_d(\text{ID}), \\
        X_I &= \operatorname{concat}(f_c, f_d).
    \end{aligned}
    \label{eqn:tokenize}
\end{equation}
Eqn.~\eqref{eqn:tokenize} proceeds as follows. Given an input image $I$, we first pass it through the vision encoder $\Phi$ to obtain continuous visual signals $f_c$. We then discretize these features using a codebook-based quantization step, producing token IDs. The token IDs are mapped to learnable ``word'' embeddings via the embedding table $E_d$ in the main MLLM, yielding discrete signals $f_d$. Finally, we concatenate $f_c$ and $f_d$ along the channel dimension to form the final visual input $X_I$ to the \textbf{Libra-1} model.
Meanwhile, a discrete auto-regressive image modeling is performed on the output features of \textbf{Libra-1}, \emph{i.e.}, each vision input is used to predict the token ID of the next position.
This simple design enables both continuous-space visual comprehension and stable discrete sequential modeling. However, since it requires the continuous signal of the entire image as input, which is not available before the image is generated, \textbf{Libra-1} is unable to perform image generation.

\begin{figure}
    \centering
    \includegraphics[width=\linewidth]{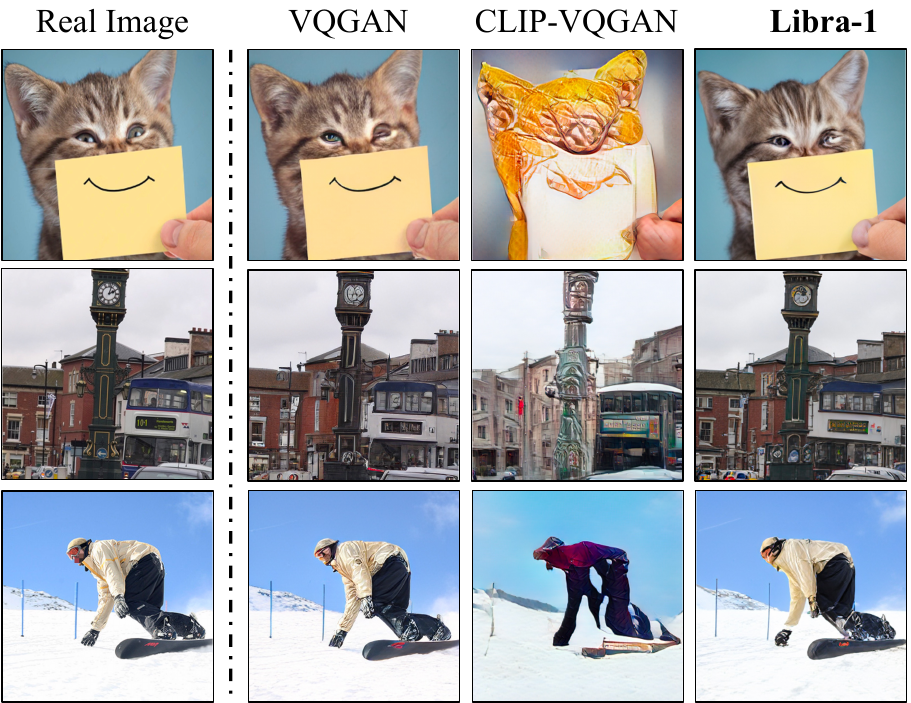}
    \caption{Image reconstruction results of the tokenizers. Directly replacing the image encoder of VQGAN with CLIP distorts the visual information. \textbf{Libra-1} largely alleviates this problem via lookup-free quantization.}
    \label{fig:clip_tokenizer}
\end{figure}

% TODO: LFQ
\textbf{Pretrained Visual Knowledge.}
To leverage the pretrained knowledge in existing well-established vision encoders like CLIP~\cite{clip}, we replace the vision encoder of VQGAN with a \textit{frozen} CLIP-ViT-L. However, training a CLIP-based VQGAN is non-trivial. The features in CLIP are highly semantic with less low-level visual information. Directly emulating such features in the quantization process of the original VQGAN results in poor reconstruction performance, as demonstrated in Fig.\,\ref{fig:cross_modal_bridge} (CLIP-VQGAN). Instead, we find that the lookup-free quantization (LFQ)~\cite{lfq}, described in Sec.~\ref{sec:lfq}, does not need to emulate the input features, largely addressing this problem. To this end, we use a CLIP-based VQGAN with LFQ as \textbf{Libra-1}'s image tokenizer. This is the first time that a highly reconstructive image tokenizer can be constructed based on a frozen vision encoder like CLIP, which has not even been investigated in the work of LFQ.

\subsection{Unified Discrete Supervision}
\label{sec:supervision}
\textbf{Libra-1} is pretrained under unified sequential modeling, where a next-token-prediction objective is performed on all input tokens., as:
\begin{equation}
p(X)=\prod_{\ell=1}^L p\left(X_{\ell} \mid X_{<\ell}\right),
\label{eqn:pretrain_loss}
\end{equation}
where $X=[X_I,X_T]$ is the input multimodal sequence and $L$ is the sequence length. In practice, the objective is computed through a discrete cross-entropy classification loss. 
However, since we introduce additional  continuous visual signals to form hybrid inputs in Sec.~\ref{sec:tokenization}, \textbf{Libra-1} is unable to generate images, as the full continuous signal is not available before the image is
generated. Therefore, the discrete autoregressive supervision on vision outputs functions solely as a regularization objective, enabling the model to learn visual world knowledge from the bottom up.

The training commonly consists of pretraining states and fine-tuning stages. The pretraining stages follows the training objective in Eqn.~\eqref{eqn:pretrain_loss}. 
The fine-tuning stages helps MLLMs to align with user intentions~\cite{rlhf,self-instruct} and generalize to unseen tasks~\cite{wei2021finetuned,chung2022scaling}. All instruction-tuning data are arranged based on this template:
\begin{equation}
    \begin{split}
        &\prompt{\text{<System Message>}}\\
        &\prompt{\text{[USER]: <Image> <Instruction>}}\\
        &\prompt{\text{[ASSISTANT]: <Answer>}}\\
    \end{split}
    \label{eqn:templete}
\end{equation}
where only \texttt{<Answer>} is accounted for computing loss, as:
\begin{equation}
p(X_{a} \mid X_{v}, X_{instruct})=\prod_{\ell=1}^L p\left(x_{\ell} \mid X_{v}, X_{instruct}, x_{<\ell}\right),
\end{equation}
where $X_{a} = \{x_{\ell}\}_{\ell=1}^{L}$, $X_v$, $X_{instruct}$ are the answers, images, and instructions.

\section{Libra-2: Towards Unified Multimodal Understanding and Generation}
\label{sec:libra2}
\subsection{Model Overview}

We further extend Libra framework to unify multimodal understanding and generation and derive \textbf{Libra-2}. The improvements are made on the following aspects:
\begin{enumerate}
    \item \textit{Architecture}. \textbf{Libra-2} also follows the core architecture design described in Sec.~\ref{sec:core_arch}. We use switch FFN to decouple multimodal understanding and generation.
    \item \textit{Representation}. Vision and language differ fundamentally in structure: images are inherently two-dimensional, whereas text is one-dimensional. To unify multimodal understanding and generation, we need a unified position encoding method for strong generation on both modalities in one model. Therefore we propose unified rotary position encoding (UniRoPE) in Sec.~\ref{sec:unirope}.
    \item \textit{Supervision}. Prior works typically employs discrete autoregression for both vision and language. To mitigate the information loss introduced by discretizing vision, \textbf{Libra-1} takes both continuous visual features and discrete IDs as hybrid inputs. This, however, disables image generation, rendering \textbf{Libra-1} an understanding-only model. In contrast, \textbf{Libra-2} implements continuous-space visual tokenization in Sec.~\ref{sec:tokenization_libra2} and continuous-space masked image generation in Sec.~\ref{sec:supervision_libra2} to enable lossless multimodal understanding and generation. 
\end{enumerate}

% Therefore, in this section, we further propose \textbf{Libra2} to extend the Libra framework to the field of unified multimodal understanding and generation. Compared to \textbf{Libra}, \textbf{Libra2} make improvements from the above three aspects. According to architecture, we propose Switch FFN in Sec.~\ref{sec:switch_ffn}. According to representation, we propose unified rotary position encoding (UniRoPE) in Sec.~\ref{sec:unirope}. According to supervision, we propose continuous-space visual tokenization in Sec.~\ref{sec:tokenization_libra2} and adaptive supervision in Sec.~\ref{sec:supervision_libra2}. 

\subsection{Unified Rotary Position Embedding}
\label{sec:unirope}

Images are inherently two-dimensional, whereas text is one-dimensional. This fundamental difference necessitates an effective position encoding strategy for unified multimodal modeling. Rotary Position Embedding (RoPE)~\cite{rope} is one of the most commonly used techniques in current MLLMs, as introduced in Sec.~\ref{sec:rope}. However, its existing variants in multimodal settings exhibit notable limitations:
\begin{enumerate}
\item Linear flattening (\eg, LLaVA~\cite{llava}): Visual features are flattened into a 1D sequence to align with the input format of language models. While this enables seamless integration, it sacrifices the intrinsic spatial structure of visual data, potentially hindering visual understanding.
\item Grouped encoding (\eg, Unified-IO~\cite{unified-io}): Attention heads are divided into two groups to encode horizontal and vertical positions separately. Although this strategy preserves spatial structure, it manually constrains the model's expressiveness: each attention head encodes only one spatial direction.
\end{enumerate}
Therefore, we propose a Unified RoPE (UniRoPE) strategy for better multimodal integration. UniRoPE splits the frequencies of RoPE into two groups for horizontal and vertical positions. Namely, for a token with position $(i,j)$, its RoPE matrix $R_{\Theta, i, j}^{d}$ is formulated as:
\begin{equation*}
\begin{adjustbox}{max width=\linewidth}
$
\setlength{\arraycolsep}{2pt}
\renewcommand{\arraystretch}{1.2}
\begin{bmatrix}
\cos(i\theta_0) & -\sin(i\theta_0) & & & & & & & \\
\sin(i\theta_0) &  \cos(i\theta_0) & & & & & & & \\
& & \cos(j\theta_1) & -\sin(j\theta_1) & & & & & \\
& & \sin(j\theta_1) &  \cos(j\theta_1) & & & & & \\
& & & & \ddots & & & & \\
& & & & & \cos(i\theta_{d/2-1}) & -\sin(i\theta_{d/2-1}) & & \\
& & & & & \sin(i\theta_{d/2-1}) &  \cos(i\theta_{d/2-1}) & & \\
& & & & & & & \cos(j\theta_{d/2}) & -\sin(j\theta_{d/2}) \\
& & & & & & & \sin(j\theta_{d/2}) &  \cos(j\theta_{d/2}) \\
\end{bmatrix}
$
\end{adjustbox}
\label{eqn:unified_rope}
\end{equation*}
where $\Theta=\left\{\theta_i=10000^{-2(i-1) / d}, i \in[1,2, \ldots, d / 2]\right\}$ is the set of predefined frequencies, and $d$ is the channel dimension. It is easy to prove that UniRoPE satisfies the relative-shift
property, as:
\begin{equation}
R_{\Theta, i_1, j_1} \cdot R_{\Theta,i_2, j_2}^\top=R_{\Theta,i_1-i_2, j_1-j_2}.
\end{equation}
In addition, in the case of text input, \emph{i.e.}, when $i = j$, UniRoPE degenerates to the standard RoPE, ensuring compatibility with existing LLMs. In fact, this is equivalent to treating each text token as an $1 \times 1$ image. 

UniRoPE enables seamless integration with existing LLM architectures while offering greater representational capacity than previous encoding strategies. 
For more detailed background on RoPE, please refer to Sec.~\ref{sec:rope} and the original paper~\cite{rope}.

% \subsection{Continuous-space Visual Tokenization}
% \label{sec:tokenization_libra2}
% \textbf{Libra2} takes fully continuous-space vision features for both multimodal understanding and generation. The input images are first encoded by a VAE~\cite{vae} to dimension 16. Then, we use a semantic augmentation module to augment the vae features with semantic features from CLIP
% \begin{equation}
% \begin{split}
%     X_I^{vae} = \Phi_{vae}(I), \\
%     X_I^{clip} = \Phi_{clip}(I), \\
%     X_I = CrossAttn(X_I^{vae}, X_I^{clip}),
% \end{split}
% \end{equation}
% where $\Phi_{vae}$ is a VAE with downsampling raito of 16 and hidden dims of 16; $\Phi_{clip}$ is a CLIP-style~\cite{clip} vision encoder, here we use SigLip-Large-384~\cite{siglip}. The cross attention takes $X_I^{vae}$ as query and $X_I^{clip}$ as key and values, with unified RoPE introduced in Sec.~\ref{sec:unirope}.
% The training is interleaved with multimodal understanding and generation samples For generation samples, we only use $X_{I}^{vae}$ as inputs for generation. For generation samples, we use the full features $X_I$ as inputs.

\subsection{Continuous-Space Visual Tokenization}
\label{sec:tokenization_libra2}

\textbf{Libra-2} operates directly on continuous-space visual features for both multimodal understanding and generation. Given an input image $I$, we first encode it with a VAE~\cite{vae}. We then semantically enrich these VAE features using a CLIP-style encoder via cross-attention:
\begin{equation}
\begin{aligned}
X_I^{\text{vae}}  &= \Phi_{\text{VAE}}(I), \\
X_I^{\text{clip}} &= \Phi_{\text{CLIP}}(I), \\
X_I              &= \operatorname{CrossAttn}(X_I^{\text{vae}},\, X_I^{\text{clip}}),
\end{aligned}
\end{equation}
where $\Phi_{\text{VAE}}$ is a VAE with downsampling factor 16 and latent channel dimension 16, and $\Phi_{\text{CLIP}}$ is a CLIP-style~\cite{clip} vision encoder (we use SigLIP-Large/384~\cite{siglip}). The cross-attention module takes $X_I^{\text{vae}}$ as queries and $X_I^{\text{clip}}$ as keys and values, using the unified RoPE introduced in Sec.~\ref{sec:unirope}.

Training interleaves multimodal understanding and image-conditioned generation examples. For generation samples, we only take $X_I^{\text{vae}}$ as input features. For understanding samples, we use the fused features $X_I$. During inference, the generated visual latent features are decoded to the pixel space by a VAE decoder.

\subsection{Adaptive Supervision}
\label{sec:supervision_libra2}
% \textbf{Libra2} uses different modeling strategies for the language and visual modalities. The language branch is supervised by auto-regression in a discrete space, while the visual branch is supervised by masked prediction in a continuous space~\cite{mar} introduced in Sec.~\ref{sec:mar}.
% We claim that continuous visual modeling is important in MLLMs, as it can enables lossless visual perception while learn vision world knowledge from the bottom up through a generative way. 
% In previous research, continuous visual perception and strong generative vision modeling are the two side of a coin that 不可兼得: 模型需要用离散的id进行高质量视觉生成~\cite{xxx}，而连续的视觉输入无法转换为合适的生成式监督信号来进行生成~\cite{}。
% 尽管有一些cascede architecure~\cite{emu}的工作尝试用自回归的方式进行连续的视觉监督，但是这样的方式很难直接生成有意义的图像，往往需要一个pretrained visual backend对预测的连续特征进一步处理才能完成生成。
% We among the first to realize continuous generative visual modeling in a unified MLLM from the bottom up.

\textbf{Libra-2} uses different modeling strategies for the language and visual modalities. The language branch is supervised by auto-regression in a discrete space, while the visual branch is supervised by masked prediction in a continuous space~\cite{mar} introduced in Sec.~\ref{sec:mar}.
We claim that continuous-space visual modeling is crucial for MLLMs: it enables lossless visual perception and supports learning visual world knowledge from the bottom up in a generative manner.
However, a key open challenge is how to equip MLLMs with strong continuous-space image generation. 
Although some cascaded architectures~\cite{emu} train auto-regressively on continuous-space visual features, the resulting models only generate an indirect prompt for a backend generator and lack explicit spatial structure.
To this end, we seek to achieve continuous, strong, and direct visual generation within a unified MLLM, enabling bottom-up learning of visual knowledge.

% suffered on a tradeoff between image generation and understanding

% directly generate meaningful images, which often requires a pretrained visual backend to post-process the predicted continuous representations.

% Prior work has treated continuous visual perception and strong generative vision modeling as mutually exclusive. Image generation within an LLM-style framework typically relies on discrete IDs for auto-regression~\cite{unified-io, lwm}, while continuous visual perception contradicts with this discretization process~\cite{llava,qwen-vl}. Although some cascade architectures~\cite{emu} attempt to use auto-regressive training over continuous vision features, the resulting models struggle to directly generate meaningful images; they often require a pretrained visual backend to post-process the predicted continuous representations.
% 
% To this end, we seek to achieve continuous, generative visual modeling within a unified MLLM, enabling bottom-up learning of visual knowledge.

We extend the continuous-space masked image generation described in Sec.~\ref{sec:mar} to the field of MLLMs. It is worth noting that this technique~\cite{mar} was  originally designed for class-to-image generation, and was later extended to text-to-image generation~\cite{fluid} . However, the mechanism of continuous-space masked image generation in unified MLLMs remains unexplored. \textbf{Libra-2} is among the first to take this step. 

Given a multimodal input comprising visual tokens $\mathbf{v} = \{v_1, v_2, \ldots, v_N\}$ and text tokens $\mathbf{t} = \{t_1, t_2, \ldots, t_M\}$, the training objective of language modeling is to maximize  its conditional probability:
\begin{equation}
P(\mathbf{t} \mid \mathbf{v}) = \prod_{i=1}^{M} p_\theta\left(t_i \mid \mathbf{v}_{<pos(t_i)},\ t_{<i}\right),
\label{eqn:ar}
\end{equation}
where $pos(x)$ denotes the position of token $x$ in the overall multimodal sequence.
Eqn.~\eqref{eqn:ar} reflects a causal attention mechanism, whereby the model predicts each text token $t_i$ based only on the preceding visual information $\mathbf{v}_{<pos(t_i)}$ and prior text tokens $t_{<i}$.

% TODO: probability
According to the visual modeling, we adopt a masked prediction strategy in the continuous space. 
Given multimodal tokens $\mathbf{v}$ and $\mathbf{t}$, the modeling process proceeds as: 1) Visual tokens are randomly replaced with a learnable [MASK] token, denoted as $v_{*}$, based on a predefined probability. 2) The model predicts the masked visual tokens based on the multimodal context. Formally,
\begin{equation}
P(\mathbf{v} \mid \mathbf{t}) = \prod_{j=1}^{N} p_\theta\left(v_{j} \mid v_*, v_2, \cdots, v_{*}, \cdots, v_{N}, \mathbf{t}_{<\text{loc}(v_j)}\right).
\label{eqn:mar}
\end{equation}
Eqn.\,\eqref{eqn:mar} indicates that the prediction for each masked token depends on all visual tokens in the image, and the preceding textual tokens. By comparing Eqns. \eqref{eqn:ar} and \eqref{eqn:mar}, it can be observed that the textual tokens still remain a causal unidirectional attention mechanism, while the visual tokens utilize a bidirectional attention mechanism in \textbf{Libra2}. This design preserves the 2D spatial structure of the visual input within the model.

% The masked visual generation strategy is adopted from MAR~\cite{mar}. It is worth noting that MAR was  originally designed for class-to-image generation, and was later extended by Fluid~\cite{fluid} to text-to-image generation. However, the mechanism of MAR in unified MLLMs remains unexplored. \textbf{Libra-2} takes the first step. 

To this end, the final loss function is defined as:
\begin{equation}
\mathcal{L} = - \alpha \log P(\mathbf{t} \mid \mathbf{v}) - \log P(\mathbf{v} \mid \mathbf{t}),
\end{equation}

During practical training,  the modeling objective of language in Eqn.~\eqref{eqn:ar} is realized by a cross-entropy (CE) loss that predict the discrete IDs of the next tokens, while the modeling objective of vision in Eqn.~\eqref{eqn:mar}  is implemented by a mean squared error (MSE) loss that predict the continuous masked visual tokens. 
We observe that the (MSE) loss on continuous visual tokens is significantly smaller than the CE loss on discrete text tokens. Therefore, we reweight the MSE loss by a factor of $\alpha = 0.25$.

%% file: sections/experiments.tex
\section{Experiments}

\subsection{Implementation Details}
% Libra + Libra2 hidden dim, layers
The cross-modal bridges in Libra architecture allow the vision and language branches to use different hidden dimensions and depths. We adopt a simple implementation: cross-modal interactions occur at every layer, and the vision branch has the same hidden size as the language branch. For \textbf{Libra-1}, we initialize the language branch with LLaMA2-7B-Chat~\cite{llama2}; for \textbf{Libra-2}, we use LLaMA3.2-1B-Instruct~\cite{llama3}. The vision branch and all cross-modal bridges are initialized from scratch. This configuration yields an 11B \textbf{Libra-1} model and a 3B \textbf{Libra-2} model. We adopt the SDXL refiner~\cite{sdxl} as a post-processing step to reduce noise in \textbf{Libra-2}'s generated images. This design does not interfere with the model’s bottom-up learning of visual knowledge: the core model is trained to reconstruct the low-level VAE latent representations, while the refiner participates neither in training nor in tokenization. serving only as post processing.

We train \textbf{Libra-1}'s image tokenizer using 10M images collected by \cite{sam}. The vision vocabulary size is enlarged to $2^{18}$ thanks to LFQ. For computational efficiency, we predict in two concatenated codebooks, each of size $2^{9}$. For the \textbf{Libra-2} image tokenizer, we employ the officially pretrained VAE~\cite{vae} and SigLiP~\cite{siglip} encoders. No further training is required.

\subsection{Training Pipeline and Data}
Training MLLMs typically comprises a multi-stage pretraining phase followed by instruction fine-tuning. We train two variants: \textbf{Libra-1}, aimed at multimodal understanding, and \textbf{Libra-2}, which unifies multimodal understanding and generation. Their training pipelines are summarized below.

\textbf{Libra-1} is trained in two stages. The training hyperparameters of \textbf{Libra-1} are shown in Tab.~\ref{tab:hyper_libra}. Below, we summarize te key pipeline:
\begin{enumerate}
    \item \textit{Pretraining}. We sample 50M image-text pairs from COYO-700M~\cite{coyo700m} and CC12M~\cite{cc12m}. To normalize caption style, we additionally include 500K image-text pairs from the COCO training split~\cite{cococap}. The data are arranged as ``\texttt{<image><text>}'' for multimodal understanding. During pretraining, the language branch is frozen; only the vision branch and the cross-modal bridges are updated.
    \item \textit{High-quality instruction fine-tuning}. We fine-tune the full model on 665K high-quality supervised samples from LLaVA-Instruct~\cite{llava}.
\end{enumerate}

\textbf{Libra-2} is trained in five stages. In Stages 1-4, the language branch is frozen, and only the vision branch and cross-modal bridges are updated. In Stage 5, we fine-tune the entire model. For all pretraining image-text pairs, we recaption them by a pretrained BLIP-2 model~\cite{blip2}. The training hyperparameters of \textbf{Libra-2} are shown in Tab.~\ref{tab:hyper_libra2}. Below, we summarize the key pipeline.

\begin{enumerate}
    \item \textit{Alignment}. We initialize the vision branch on ImageNet-1K~\cite{imagenet} to learn basic pixel dependencies, arranging the data in a class-to-image generation format, namely, ``\texttt{<name><image>}'', where \texttt{<name>} represents the text names of each class.
    \item \textit{Pretraining (PT)}. This is the main training stage. We train on 200M image-text pairs from LAION-COCO~\cite{laion-coco}, interleaving two sequence formats: ``\texttt{<image><text>}'' for multimodal understanding and ``\texttt{<text><image>}'' for multimodal generation. As understanding is easier to learn than generation, we upweight the latter by sampling the two formats with probabilities of 20\% (understanding) and 80\% (generation).
    \item \emph{Optional: Resolution scaling.} Higher input resolutions generally improve multimodal performance. We therefore optionally increase the training resolution on LAION--COCO from $256\times256$ to $384\times384$ pixels. If the resolution is scaled at this stage, the same scaled resolution should be used in all subsequent steps.
    \item \textit{Aesthetic modulation (CT)}. As images from LAION-COCO contain massive low-quality visual renderings, we further modulate \textbf{Libra-2}'s image generation toward higher aesthetic quality. We use 14M aesthetic-oriented samoles from JourneyDB~\cite{journeydb} and LAION-Aesthetic~\cite{laion-aesthetic}, using the same data arrangement as in Stage~2.
    \item \textit{High-quality instruction fine-tuning (SFT)}. We fine-tune the full model on the 1.2M high-quality supervised samples from InternVL~\cite{internvl} for multimodal understanding and on 1M samples drawn from JourneyDB and LAION-Aesthetic to maintain the multimodal generation ability. 
\end{enumerate}

\input{tables/hyperparam_libra}
\input{tables/hyperparam}

\input{tables/vqa}

\input{tables/geneval}

\subsection{Multimodal Understanding}
% vqa, caption, mllm
% instance: understanding & generation

% We evaluate Libra-series models' multimodal understanding ability on a wide range of benchmarks, including traditional VQA, image captioning, and general MLLM benchmarks. Tab.~\ref{tab:vqa} shows the results. 
% We first compare our understanding-only model, \textbf{Libra}, with previous works. The results confirm that \textbf{Libra} rivals existing modern MLLMs on multimodal understanding.
% We further make systemetic comparison of under the field of unified multomodal undertsanding and generation. Several observations: 1) Libra architecture consistently outperform unified and cascade architectures. Compared to Show-O, which Our data usage is most closed to the one in it, Libra largely outperform it on all benchmarks. 2) Higher resolution is important for accurate multimodal understanding. When we scale Libra2's input resolution from 256 to 384, performance largely improve. 
% Also, Compare Libra2 with Libra, despite Libra2 has much less parameters, it can achieve compatible results on several benchmarks, such as VQAv2, MME, POPE, SEED. This suggests that multimodal Generation can help multimodal understanding.
% We also observe that \textbf{Libra-2} perform not good at image captioning. This may because the training data we use are mostly with generated long captions, but coco only consider a short sentence. Such a distribution difference on the text may result in model not to summarize all the key part in a short sentence.

\textbf{Traditional understanding benchmarks}. We evaluate the multimodal understanding capabilities of the Libra series across a broad range of benchmarks, including traditional VQA and image captioning. Tab.~\ref{tab:vqa} summarizes the results. We first compare the understanding-only model, \textbf{Libra-1}, with prior work and find that it surpasses contemporary MLLMs on multimodal understanding. We also observed that \textbf{Libra-2} underperforms on image captioning. This probably stems from a distribution mismatch: our training data predominantly contain long captions, whereas COCO emphasizes concise, single-sentence captions. This mismatch discourages the model from compressing key content into short summaries.

\textbf{General-purpose MLLM benchmarks.}
We then conduct a systematic comparison in the field of unified multimodal understanding and generation. The main observations are:
\begin{enumerate}
    \item The Libra architecture consistently outperforms unified and cascaded architectures. Compared with Show-O, whose data scale and composition most closely match ours, \textbf{Libra-2} achieves substantially higher scores across all benchmarks.
    \item Higher input resolution is important for accurate multimodal understanding. Increasing \textbf{Libra-2}'s resolution from $256^2$ to $384^2$ yields significant performance gains.
    \item Generation helps understanding. Despite having far fewer parameters than \textbf{Libra-1}, \textbf{Libra-2} attains comparable results on several benchmarks (e.g., VQAv2, MME, POPE, SEED), suggesting that multimodal generation can benefit multimodal understanding.
\end{enumerate}

% Compared to the previous best unified model, Janus~\cite{janus}, Libra2 xxxx, with much simpler and fully open-sourced training data. 
% Compared to Libra, Libra2 xxx.
% Several interesting observations can be found in the results in Tab.\,\ref{tab:vqa}. (1) Architecture. (2) Generation enhance understanding. (3) Resolution.

% First, architecture
% Second, CLIP-based semantic alignment
% Third, resolution.

% We evaluate \textbf{Libra-2}'s multimodal understanding on a wide range of benchmarks, including traditional VQA, image captioning, and general MLLM benchmarks. Compared to the previous best unified model, Janus~\cite{janus}, Libra2 xxxx, with much simpler and fully open-sourced training data. 
% Compared to Libra, Libra2 xxx.
% Several interesting observations can be found in the results in Tab.\,\ref{tab:vqa}. (1) Architecture. (2) Generation enhance understanding. (3) Resolution.

% We evaluate \textbf{Libra-2}'s multimodal understanding on a wide range of benchmarks, including traditional VQA, image captioning, and general MLLM benchmarks. Compared to the previous best unified model, Janus~\cite{janus}, Libra2 xxxx, with much simpler and fully open-sourced training data. 
% Compared to Libra, Libra2 xxx.
% Several interesting observations can be found in the results in Tab.\,\ref{tab:vqa}. (1) Architecture. (2) Generation enhance understanding. (3) Resolution.

\begin{figure*}[t]
    \centering
    \includegraphics[width=\linewidth]{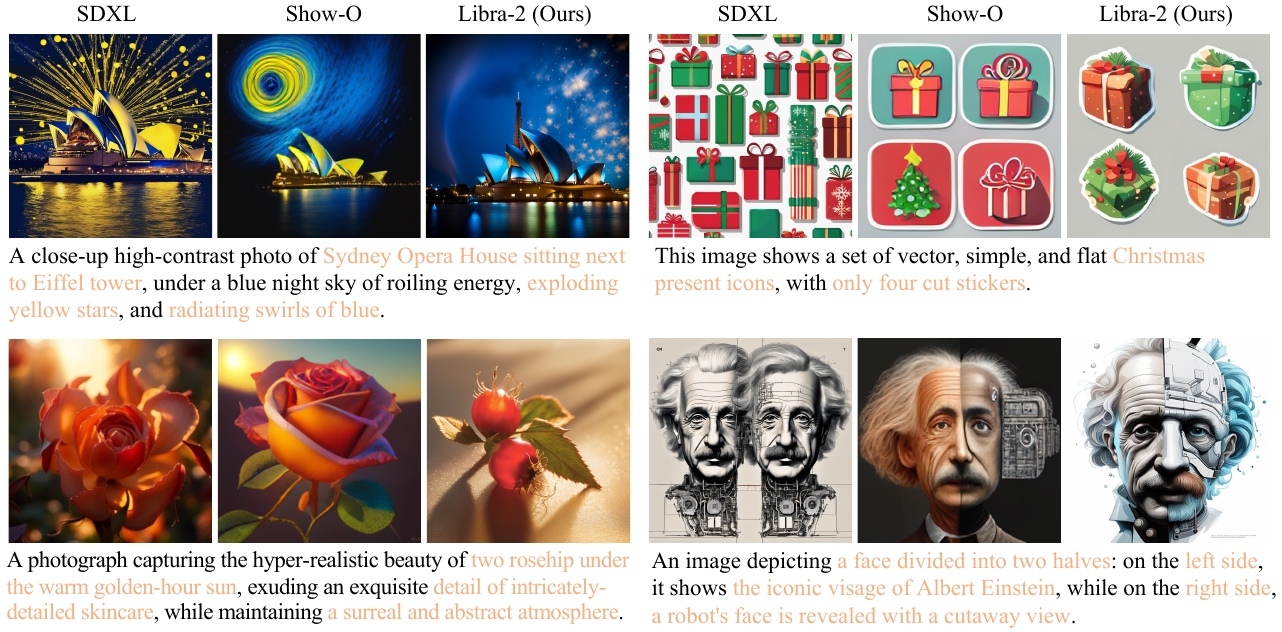}
    \caption{Comparison on text-to-image generation. We highlight the key parts of the text prompts, where \textbf{Libra-2} demonstrates strong prompt-following capacity.
    % shows several text-to-image generation results of \textbf{Libra-2}. More results are presented in Sec.~\ref{}. In Fig~\cite{fig:t2i}, we compare the text-to-image generation results of \textbf{Libra-2} with two baselines, StableDiffusionXL~\cite{sdxl} and Show-O~\cite{show-o}, where StableDiffusionXL is a widely-used strong text-to-image generation model, and Show-O is a unified MLLM that use the most closed training
    }
    \label{fig:t2i}
\end{figure*}

\subsection{Text-to-Image Generation}
% coco fid, mjhd fid
% geneval

\input{tables/fid}

\textbf{General ability of text-to-image generation}.
A key metric for text-to-image generation is a model’s ability to follow textual instructions. We evaluate \textbf{Libra-2}’s general generative capability using the widely adopted GenEval~\cite{geneval} benchmark. As shown in Tab~\ref{tab:geneval}, \textbf{Libra-2} ranks among the leading models and even surpasses systems designed specifically for generation tasks. 
Notably, \textbf{Libra-2} excels at ``counting,'' a long-standing weakness of prior models, outperforming the previous state of the art by 12\%. This may because counting is closely tied to language comprehension: the model must interpret numerals in text and map them to their visual realizations. Prior approaches may not have captured this alignment effectively. By enabling bottom-up alignment between visual and textual modalities, the Libra architecture affords a more faithful grounding of numeric concepts.

\textbf{Visual quality of image generation}. 
We evaluate \textbf{Libra-2}’s image generation quality using the FID metric~\cite{fid}, as shown in Tab.\,\ref{tab:fid}. Two benchmarks are used, where COCO~\cite{coco} focuses on realism and MJHQ~\cite{mjhq} emphasizes aesthetics. Compared to other unified models, \textbf{Libra-2} consistently produces high-quality images, achieving the best FID score.
To quantify the improvement of \textbf{Libra-2} over \textbf{Libra-1}, we build an image-generation-capable variant of \textbf{Libra-1} by disabling the contiguous visual signal (see Sec.~\ref{sec:supervision}) and fine-tuning it on an additional 10M image-text pairs from the pretraining data. \textbf{Libra-2} substantially mitigates \textbf{Libra-1}'s shortcomings in image generation, achieving a relative FID reduction of over 80\% on both COCO and MJHQ.

It is worth noting that FID is a vision-centric metric, focusing solely on how closely the generated images resemble the target images, thus primarily reflecting the effectiveness of vision modeling, rather than cross-modal comprehension.
Based on this, we attribute \textbf{Libra-2}’s superior performance on FID to its separated vision branch, which models visual information independently. In contrast, previous works typically couple vision and language modeling within a single backbone, potentially reduce the focus on vision. By separating the vision branch, \textbf{Libra-2} allows for more specialized and effective vision modeling, reducing the impact of language modeling on vision.

% \begin{figure}[t]
%     \centering
%     \includegraphics[width=\linewidth]{images/sampling_steps.pdf}
%     \caption{Sampling steps.}
%     \label{fig:sample_steps}
% \end{figure}

% \begin{figure*}[t]
%     \centering
%     \includegraphics[width=\linewidth]{images/t2i_image_backup.pdf}
%     \caption{Comparison on text-to-image generation. We highlight the key parts of the text prompts, where \textbf{Libra-2} demonstrates strong prompt-following capacity.
%     % shows several text-to-image generation results of \textbf{Libra-2}. More results are presented in Sec.~\ref{}. In Fig~\cite{fig:t2i}, we compare the text-to-image generation results of \textbf{Libra-2} with two baselines, StableDiffusionXL~\cite{sdxl} and Show-O~\cite{show-o}, where StableDiffusionXL is a widely-used strong text-to-image generation model, and Show-O is a unified MLLM that use the most closed training
%     }
%     \label{fig:t2i}
% \end{figure*}

\textbf{Case study}.
% Fig.~\ref{fig:t2i} shows several text-to-image generation results of \textbf{Libra-2}. More results are presented in Sec.~\ref{}. In Fig~\cite{fig:t2i}, we compare the text-to-image generation results of \textbf{Libra-2} with two baselines, StableDiffusionXL~\cite{sdxl} and Show-O~\cite{show-o}, where StableDiffusionXL is a widely-used strong text-to-image generation model, and Show-O is a unified MLLM that use the most closed training data with ours. We can see that \textbf{Libra-2} have a more deep understanding about the text prompt. Such as ``Sydney Opera House sitting next to Eiffel tower''. We can also see that \textbf{Libra-2} can accurately generate the given number of objects, like ``two rosehip'' and ``four cut stikers''.
% 
Fig.~\ref{fig:t2i} showcases representative text-to-image results from \textbf{Libra-2}. We compare \textbf{Libra-2} with two baselines—Stable Diffusion XL (SDXL)~\cite{sdxl}, a widely used state-of-the-art text-to-image model, and Show-O~\cite{show-o}, a unified MLLM trained on data most similar to ours. As shown, \textbf{Libra-2} demonstrates a deeper understanding of complex prompts (\emph{e.g.}, ``Sydney Opera House next to the Eiffel Tower'') and more precise control over object counts, accurately rendering the specified quantities (\emph{e.g.}, ``two rosehips'' and ``four cut stickers'').

% \textcolor{blue}{We show several failure cases in Fig.~\ref{fig:failure_gen}, where we can observe several types of short-coming of \textbf{Libra-2}'s image generation. First, the visual detail not enough. As seen the clock in the first image do not have clear 刻度. 这可能是由于生成图像的分辨率和vae编码器受限。Second, it is hard to generate accurate and clear words. As in the second image, The word ``Libra'' is not correctly generated. 这可能是因为训练数据没有文本生成，模型只能从零散的噪声数据中学习，故而没有很鲁棒的将文字和视觉对应起来。Third, it can not well solve the position relationship. As in the third image, if the model do not understand the location ``right of'', it perfer to 组合 the two things into one thing. 这可能是因为训练数据大多是都是以单个物体存在，并没有明确的位置引导。

\begin{figure}[t]
    \centering
    \includegraphics[width=\linewidth]{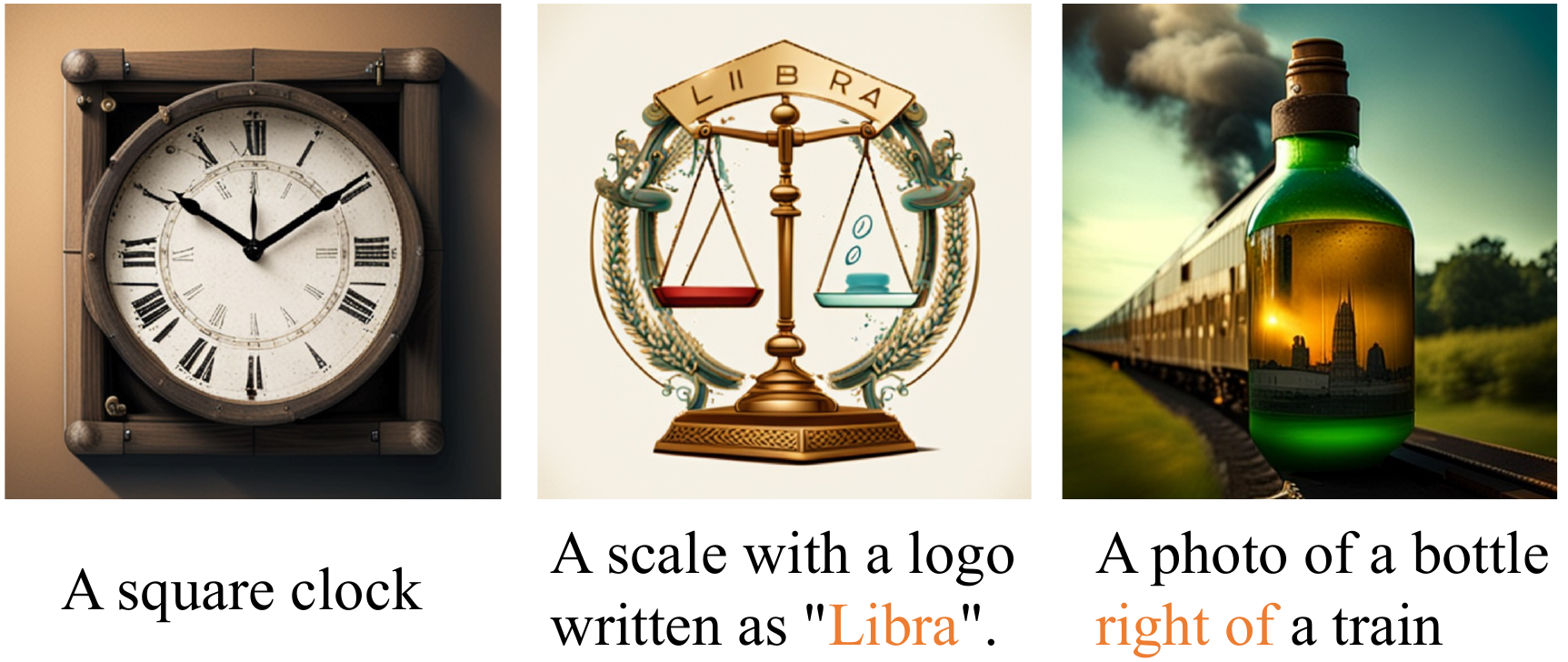}
    \caption{Several failure cases of \textbf{Libra-2}'s generated images.}
    \label{fig:failure_gen}
\end{figure}

We present several failure cases in Fig.~\ref{fig:failure_gen}, which illustrate multiple shortcomings of \textbf{Libra-2}'s image generation. First, the visual detail is insufficient: for instance, the clock in the first image lacks clear tick marks. This may be due to the limited resolution of the generated images and constraints imposed by the VAE encoder. Second, the model struggles to render accurate, legible text; in the second image, the word ``Libra'' is not correctly produced. 
A plausible explanation is that textual content is highly structured, which is friendly to perception (\emph{e.g.}, OCR~\cite{ocr}) but correspondingly raising difficulties in generation: even small errors can cause large readability degradations. Achieving robust text generation typically requires a lot of specialized data to strengthen the model’s text-generation capability. However, the Libra series is trained on noisy internet image-text pairs with only sparse textual signals, which hinders reliable alignment between words and visual content.
Third, the model often fails to resolve spatial relations. In the third image, when it does not properly understand the relation ``to the right of,'' it tends to merge the two objects into one. This may stem from training data that mostly feature single objects and lack explicit positional supervision.

\subsection{Analysis on Understanding-only Modeling}
% In this section, we ablate the basic architecture design of Libra-series architecture through \textbf{Libra} model. 
In this section, we present ablation studies of the core architectural design of the Libra series, using the \textbf{Libra-1} model as a representative case. The evaluation mainly focuses on three general multimodal understanding benchmarks: MME~\cite{mme}, POPE~\cite{pope}, and SEED~\cite{seed}.

\textbf{Libra architecture v.s Unified architecture.}
In Tab.\,\ref{tab:ablation_libra}(a), we remove the switch attention design in Libra, where Libra degenerates to a unified architecture like Show-O~\cite{show-o}. The results show that the coupled vision and language systems exhibit obvious performance degradation on zero-shot tasks.

We further investigate the impact of the cross-modal bridge within the switch-attention module, as shown in Tab.~\ref{tab:ablation_libra}(b). The results indicate that adding only an additional vision branch without cross-modal bridge yields limited gains and consistently performs worse than the original \textbf{Libra-1} model. Moreover, compared to the unified-architecture variant (Tab.~\ref{tab:ablation_libra}(a)), the variant without the cross-modal bridge (Tab.~\ref{tab:ablation_libra}(b)) shows limited improvement, despite its larger parameter size. his indicates the importance of a reasonable cross-modal interaction strategy. Fig,~\ref{fig:cross_modal_bridge} provides additional evidence of the effectiveness of the cross-modal bridge.

\input{tables/ablation_libra}

\textbf{Impact of hybrid vision inputs.}
In \textbf{Libra-1}, we take the contiguous visual signal along with the discrete tokenized embeddings as hybrid inputs for both accurate visual perception and robust generative visual modeling.
We claim that the contiguous visual signal plays a crucial role in accurate visual perception. To validate this, we remove the contiguous visual signals in \textbf{Libra-1}'s hybrid inputs and only retain  the discrete embeddings. As shown in Tab.\,\ref{tab:ablation_libra}(c), a clear performance degradation rises when solely using discrete inputs (\emph{e.g,}, -14.2\% on SEED). 

\textbf{Impact of unified supervision}.
\textbf{Libra-1} employs unified discrete autoregressive modeling across both the language and vision modalities. Although it cannot generate images due to the hybrid visual inputs, the vision-side modeling enables the model to acquire visual knowledge in a bottom-up, generative manner. Tab.~\ref{tab:ablation_libra}(d) shows a variant trained with supervision only on the language side, which yields a consistent drop in performance. This indicates the importance of vision modeling for in-depth visual understanding.

% \textbf{Libra} uses a unified discrete auto-regressive modeling on both language and vision sides. Despite it is unable to generate images duo to the hybrid visual inputs, the modeling on the vision side helps the model learn visual knowledge from the bottom up in a generative way. Tab.~\ref{tab:ablation_libra}(d) shows a variant that only supervise the language side. It can be seen that the performance consistently drop. This indicate that the importance of vision modeling that help the model understand the visual world better.

% \begin{figure}[t]
%     \centering
%     \includegraphics[width=\linewidth]{images/attention.pdf}
%     \caption{Failure cases.}
%     \label{fig:arch_comp}
% \end{figure}

\begin{figure}[t]
\centering
\subfloat[$D_{l}^{layer}$]{\includegraphics[width=0.5\columnwidth]{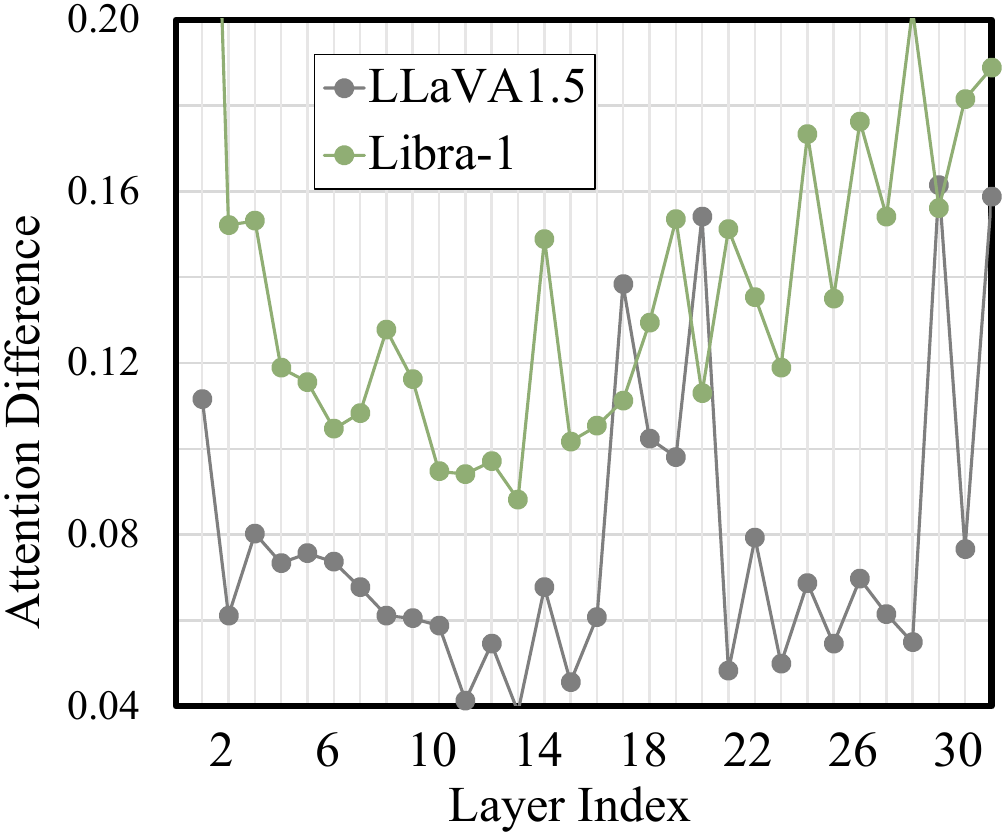}}
\hfill
\subfloat[$D_{l}^{head}$]{\includegraphics[width=0.5\columnwidth]{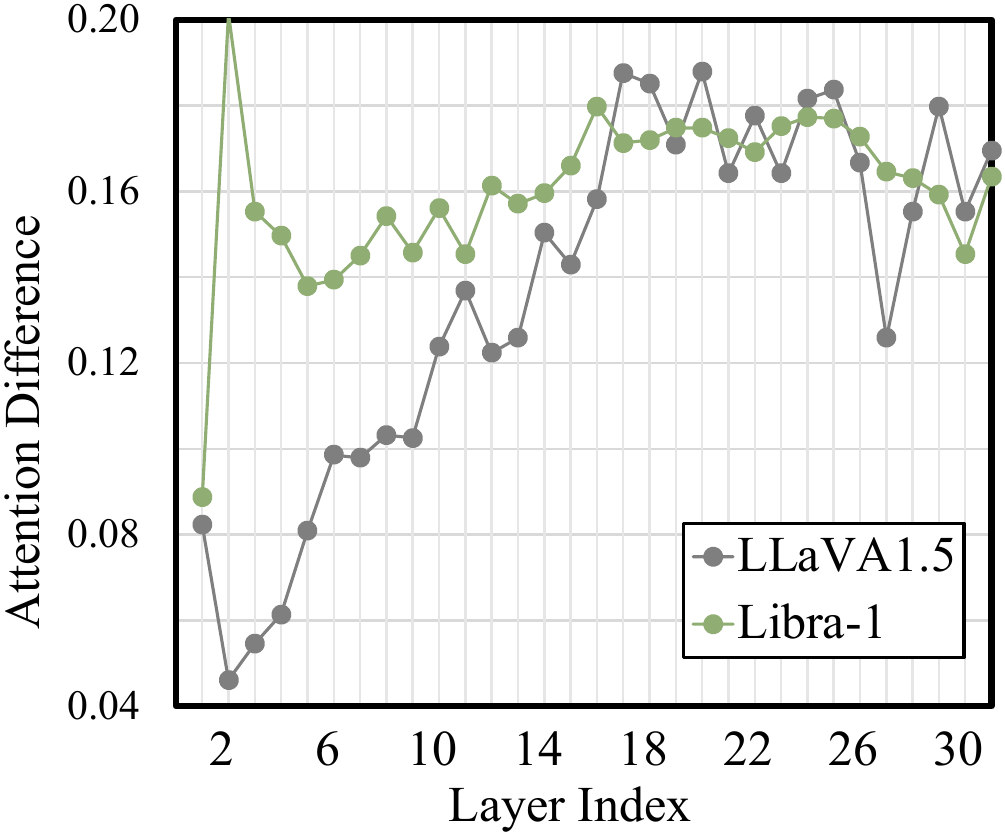}}
\caption{Mean absolute error (MAE) of the attention activations. (a) Layer-wise attention diversity. (b) Head-wise attention diversity.}
\label{fig:vis_attn}
\end{figure}

% \begin{figure}[t]
% \centering
% \begin{subfigure}{0.50\columnwidth}
% \includegraphics[width=\linewidth]{images/attention.pdf}
% \caption{}
% \end{subfigure}
% \hspace{-5pt}
% \begin{subfigure}{0.50\columnwidth}
% \includegraphics[width=\linewidth]{images/attention.pdf}
% \caption{}
% \end{subfigure}
% \hspace{-5pt}
% \begin{subfigure}{0.50\columnwidth}
% \includegraphics[width=\linewidth]{images/attention.pdf}
% \caption{}
% \end{subfigure}
% \hspace{-5pt}
% \begin{subfigure}{0.50\columnwidth}
% \includegraphics[width=\linewidth]{images/attention.pdf}
% \caption{}
% \end{subfigure}
% \vspace{-20pt}
% % \captionsetup{font=small}   
% \caption{xxx}
% \label{fig:ablation_epoch}
% \vspace{-15pt}
% \end{figure}

% \begin{figure}[t]
% \centering
% \subfloat[]{\includegraphics[width=\columnwidth]{images/attention.pdf}}\\[2pt]
% \hfill
% \subfloat[]{\includegraphics[width=\columnwidth]{images/attention.pdf}}
% \caption{xxx}
% \label{fig:ablation_epoch}
% \end{figure}

\textbf{Attention diversity}.
We analyze \textbf{Libra-1}'s attention patterns on a VQA task in Fig.~\ref{fig:vis_attn}, averaging results over 100 VQA samples. We report two metrics based on the mean absolute error:
\begin{enumerate}
    \item \textit{Layer-wise attention diversity}. In Fig.~\ref{fig:vis_attn}(a), we measure how each layer’s answer-to-image attention differs from the cross-layer mean. For a sample $s$, let $A_{l,h}^{(s)} \in \mathbb{R}^{N}$ denote the attention activation over $N$ image tokens induced by the answer token at layer $l$ and head $h$. Define the layer-averaged activation
    $
    A_{l}^{(s)} = \frac{1}{H} \sum_{h=1}^{H} A_{l,h}^{(s)}
    $
    and the cross-layer mean
    $
    \bar{A}^{(s)} = \frac{1}{L} \sum_{l=1}^{L} A_{l}^{(s)}.
    $
    The layer-wise attention difference is
    $$
    D^{\text{layer}}_{l} = \frac{1}{S} \sum_{s=1}^{S} \frac{1}{N} \left\lVert A_{l}^{(s)} - \bar{A}^{(s)} \right\rVert_{1},
    $$
    where $S = 100$. Larger values indicate greater across-layer diversity.
    \item \textit{Head-wise attention diversity}. In Fig.~\ref{fig:vis_attn}(b), we quantify the diversity across heads within each layer. For sample $s$ and layer $l$, let
    $
    \bar{A}_{l}^{(s)} = \frac{1}{H} \sum_{h=1}^{H} A_{l,h}^{(s)}.
    $
    The head-wise deviation for layer $l$ is
    $$
    D^{\text{head}}_{l} = \frac{1}{S} \sum_{s=1}^{S} \frac{1}{H} \sum_{h=1}^{H} \frac{1}{N} \left\lVert A_{l,h}^{(s)} - \bar{A}_{l}^{(s)} \right\rVert_{1}.
    $$
    Larger values indicate greater across-head diversity within that layer.
\end{enumerate}

Fig.~\ref{fig:vis_attn} reveals several interesting findings: 1) LLaVA1.5~\cite{llava1.5}, a well-known MLLM baseline, demonstrates similar attention patterns across layers, with lower attention diversity, while \textbf{Libra-1} shows diverse attention patterns across all layers. These results suggest that the decoupled vision-language system has lower learning redundancy in both layer-wise and head-wise aspects, as observed through diverse attention patterns.

\input{tables/ablation}

\subsection{Analysis on Unified Modeling}

In \textbf{Libra-2}, we extend the Libra architecture to unified multimodal understanding and generation. We introduce important improvements including Switch-FFN, UniRoPE, and continuous visual modeling, focusing on architecture, representation, and supervision, respectively.
Below, we analyze the impact of these improvements. We evaluate on one understanding benchmark, MME~\cite{mme}, and one generation benchmark, GenEval~\cite{geneval}.

\textbf{Impact of Switch FFN}.
In Sec.~\ref{sec:arch}, we incorporate a Switch FFN into \textbf{Libra-2} to decouple understanding from generation. Tab.~\ref{tab:ablation}(a) presents a variant of \textbf{Libra-2} in which the Switch FFN is replaced with a vanilla FFN. Removing the Switch FFN decreases MME by 69 points and GenEval by 2\%. We attribute this degradation to a granularity mismatch between understanding and generation: a single vanilla FFN forces both into a shared feature space, resulting in inferior performance.

\textbf{Impact of unified RoPE}.
As discussed in Sec.~\ref{sec:unirope}, vision and language exhibit a key structural difference: visual data are organized in two-dimensional spatial layouts, whereas language is arranged as a one-dimensional causal sequence. We therefore propose UniRoPE, a unified positional encoding that reconciles these two modalities. This design is crucial for high-quality cross-modal understanding and generation. As shown in Tab.~\ref{tab:ablation}(b), replacing UniRoPE with vanilla RoPE in \textbf{Libra-2} substantially reduces the MME score from 1475 to 1252 and the GenEval score from 0.63 to 0.42, which largely support the effectiveness of UniRoPE.

% \textbf{Impact of Continuous Image Modeling}.
% In \textbf{Libra}, we use a hybrid tokenization strategy to make the model take both continuous and discrete signals as inputs and supervise it with discrete autoregression. The hybrid tokenization enables model to receive lossless continuous visual features, while performing discrete modeling to make the model in-depth understand the data. But this strategy can not use as generation as during generation we can not access the whole continuous signals of the generated images before it is generated.

% To overcome this, in \textbf{Libra-2}, we use masked image generation in continuous space, which directly make the model take continuous image features as input, and autoregressively predict the continuous features as supervision. This is important for MLLMs. As multimodal comprehension requires finegrained visual information. 

% Tab.~\ref{}(c) make comprison on these two type of tokenization by replace the inputs of \textbf{Libra-2} with hybrid inputs. We can see in this way model can not generate images, and have a performance degradation on multimodal understanding (-59 on MME). This suggests that the masked image generation in continuous space we used in \textbf{Libra-2} not only enables high quality image generation, but also further improve the understanding.

\textbf{Impact of continuous image modeling}.
In \textbf{Libra-1}, we adopt a hybrid tokenization strategy that allows the model to take both continuous and discrete signals as inputs while being supervised with discrete autoregression. This hybrid design lets the model access lossless continuous visual features while performing discrete modeling that promotes deeper data understanding; however, it cannot be used for generation, because the whole continuous signals of an image are unavailable before the image itself is generated. To address this, \textbf{Libra-2} employs masked image generation in continuous space, feeding continuous image features directly to the model and training it with a masked generation manner. The continuous vision space is crucial for MLLMs, where multimodal understanding relies on fine-grained visual information.

Tab.~\ref{tab:ablation}(c) compares these two tokenization strategies by replacing the \textbf{Libra-2}'s inputs with hybrid tokens and training the model, as in \textbf{Libra-1}, under a unified discrete autoregressive manner for both the vision and language modalities. Under this setting, the model cannot generate images and exhibits an obvious degradation in multimodal understanding (-59 on the MME score). This is because the hybrid tokenization strategy in \textbf{Libra-1} introduces discrete noise into the continuous features during the hybridization process. \textbf{Libra-2}'s continuous-space masked image generation not only enables high-quality image generation but also strengthens understanding.

\input{tables/ablation_obj}

\noindent\textbf{Mutual impact of generation and understanding.}
Tab.~\ref{tab:gen_vs_und} compares the unified \textbf{Libra-2} (Und.~\&~Gen.) model with its single-task variants: an understanding-only model and a generation-only model. The single-task variants are obtained by adjusting the sampling probabilities of understanding and generation data during training (see ``T2I:I2T ratio'' in Tab.~\ref{tab:hyper_libra2}). We construct understanding and generation samples from the same image-text pairs by swapping the order of modalities in the sequence, \emph{i.e.}, ``\texttt{<image><text>}'' for understanding and ``\texttt{<text><image>}'' for generation. During instruction tuning, we directly control the ratio of understanding to generation data; the understanding-only and generation-only variants simply use one type of data exclusively.

The results in Tab.~\ref{tab:gen_vs_und} indicate that adding generation training has a modest yet generally positive effect on understanding. Compared with the understanding-only variant, the unified \textbf{Libra-2} improves VQAv2 from 76.0\% to 77.5\% and MMB from 58.2\% to 60.0\%. In the opposite direction, understanding training also benefits generation: compared with the generation-only variant, the unified \textbf{Libra-2} consistently achieves better performance across all three generation benchmarks. Overall, understanding and generation are mutually reinforcing.

% \begin{figure}[t]
% \centering
% \subfloat[]{\includegraphics[width=0.5\columnwidth]{images/mask_sampling_steps.pdf}}
% \hfill
% \subfloat[]{\includegraphics[width=0.5\columnwidth]{images/diffusion_sampling_steps.pdf}}
% \caption{GenEval performance versus sampling steps for multiple CFG scales: (a) masking sampling step, (b) diffusion sampling step.}
% \label{fig:sample_steps}
% \end{figure}

\begin{figure}[t]
    \centering
    \includegraphics[width=\columnwidth]{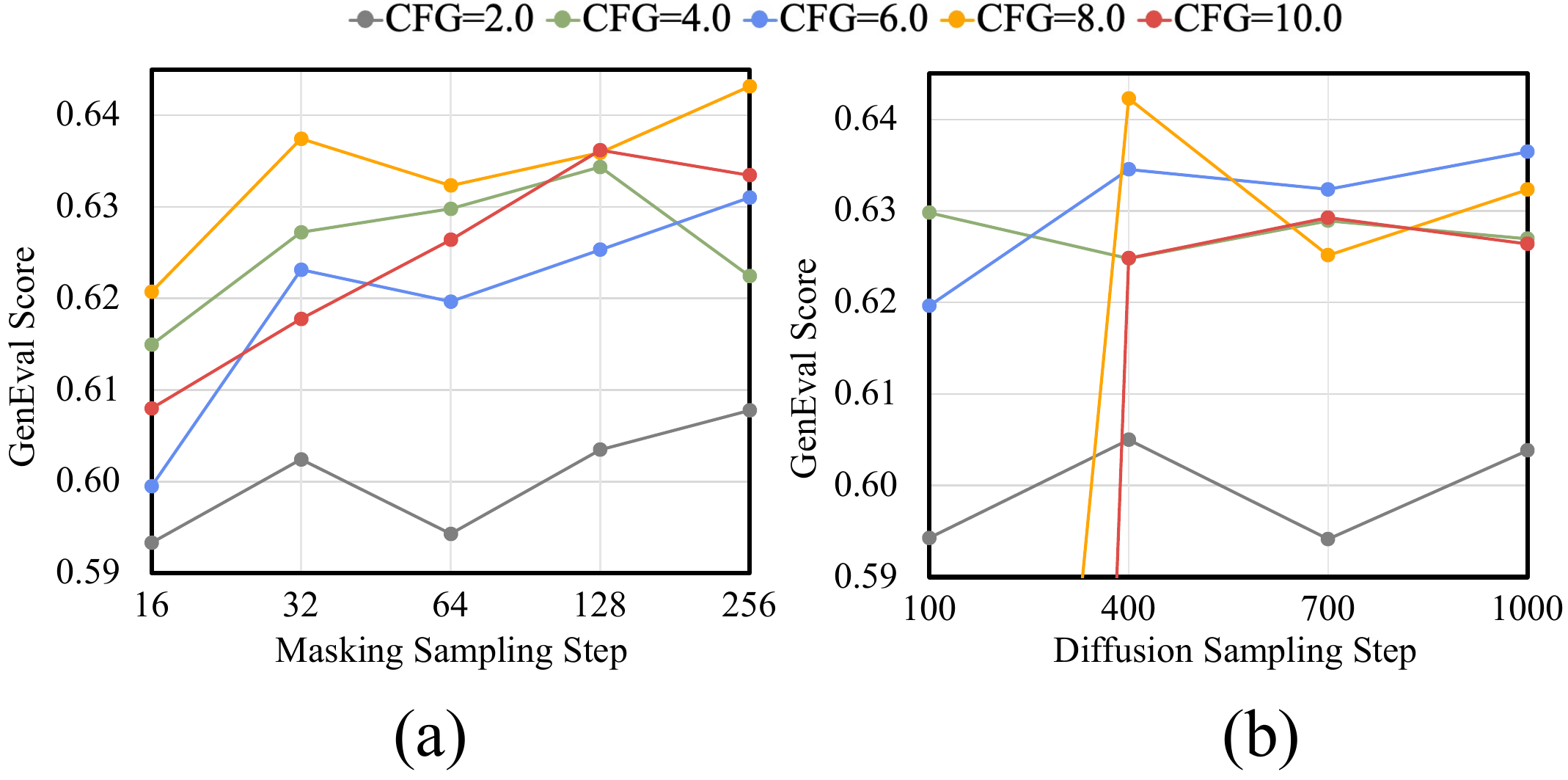}
    \caption{GenEval performance versus sampling steps for multiple CFG scales: (a) masking sampling step, (b) diffusion sampling step.}
    \label{fig:sample_steps}
\end{figure}

% \begin{figure}[t]
% \centering
% \subfloat[]{\includegraphics[width=\columnwidth]{images/sampling_steps.pdf}}\\[2pt]
% \hfill
% \subfloat[]{\includegraphics[width=\columnwidth]{images/attention.pdf}}
% \caption{xxx}
% \label{fig:ablation_epoch}
% \end{figure}

% \textbf{Impact of Sampling steps.}
% There are two type of sampling steps in \textbf{Libra-2}: 1) the masking sampling step, which the masked image generation is a autoregression-like process that progressively generate image tokens in several steps. and the masking sampling step controls the sample 精细程度 when generating the whole image.
% 2) the diffusion step in the lightweight diffusion head when predicting each image token. the diffusion step controls the 精细程度 when generating a single token.
% We test the GenEval score on different CFG scales~\cite{cfg}. 
% In the left figure of Fig.~\ref{}, we can see that the generation ability increase as the masking sampling step increase.
% In the right figure of Fig.~\ref{}, we can see that diffusion step do not have very much impact on the final generation capacity. This is because we do not use diffusion to directly generate a whole image, but just  one token. This task is easy that just a low sampling step number is sufficient. There is a great performance degradation on high CFG scale (cfg=8.0 and 10.0) when the diffusion step is below 400. This is because a high CFG scale always bring more sampling difficulty, where when diffusion step below 400 the model is easily to accumulate the noise so that can not generate meaningful images.

\textbf{Impact of sampling steps.}
We consider two types of sampling steps in \textbf{Libra-2}: (1) masking sampling steps, where masked image generation is an autoregressive-like process that progressively generates image tokens over multiple steps; the number of masking steps controls the granularity of whole-image generation; and (2) diffusion steps in the lightweight diffusion head used to predict each image token; the number of diffusion steps controls the token-level granularity.

We evaluate GenEval across different classifier-free guidance (CFG) scales~\cite{cfg}. As shown in Fig.~\ref{fig:sample_steps}(a), increasing the number of masking sampling steps consistently improves generation quality. In contrast, Fig.~\ref{fig:sample_steps}(b) shows that the number of diffusion steps has only a modest effect on overall performance. This is because diffusion is applied only to individual tokens rather than to the entire image, so a relatively small number of steps is sufficient for this easier subtask. 
Meanwhile, higher CFG scales generally yield better generations. However, at high CFG (\emph{e.g.},~8.0 and 10.0) combined with a low number diffusion steps, performance degrades substantially: when the number of diffusion steps is below 400, stronger guidance makes sampling more brittle, and too few sampling steps cause accumulated noise, resulting in meaningless images.

%% file: tables/hyperparam_libra.tex
\begin{table}[t]
\caption{Training hyperparameters of \textbf{Libra-1}.}
\label{tab:hyper_libra}
\vskip 0.1in
\begin{center}
\begin{small}
% \begin{sc}
% \resizebox{\linewidth}{!}{
% \setlength{\tabcolsep}{1mm}{
\begin{tabular}{l|ll}
\toprule
Configuration       & Pretraining        & SFT        \\
\midrule
Total steps         & 40000              & 7000       \\
Warmup steps        & 2000               & 300        \\
Batch size          & 1280               & 128        \\
Learning rate       & 1e-4               & 2e-5       \\
Learning rate decay & \multicolumn{2}{c}{cosine decay}      \\
Weight decay        & \multicolumn{2}{c}{0.01}        \\
Dropout ratio       & \multicolumn{2}{c}{0.0}         \\
Optimizer           & \multicolumn{2}{c}{AdamW}       \\
Adam $\epsilon$              & \multicolumn{2}{c}{1e-8}        \\
Adam $\beta$              & \multicolumn{2}{c}{(0.9, 0.99)} \\
Gradient clipping   & \multicolumn{2}{c}{1.0}         \\
Numerical precision & \multicolumn{2}{c}{bfloat16}    \\ 
\midrule
LLM                 & \multicolumn{2}{c}{LLaMA2-7B-Chat}        \\
Vision encoder      & \multicolumn{2}{c}{CLIP-ViT-L-336px}        \\
Image resolution    & \multicolumn{2}{c}{336$^{2}$}         \\
Patch size          & \multicolumn{2}{c}{14 $\times$ 14}         \\
Image token number    &  \multicolumn{2}{c}{578}              \\
Vision vocab size   & \multicolumn{2}{c}{$(2^{9})^{2} = 2^{18}$} \\
Language vocab size & \multicolumn{2}{c}{32000} \\
\bottomrule
\end{tabular}
% }
% }
% \end{sc}
\end{small}
\end{center}
\vskip -0.1in
\end{table}

%% file: tables/hyperparam.tex
\begin{table}[t]
\centering
\caption{Training hyperparameters of \textbf{Libra-2}.}
\label{tab:hyper_libra2}
\begin{tabular}{l|cccc}
\toprule
Configuration        & Alignment            & PT                & CT                & SFT               \\
\midrule
Learning rate        & 1e-4                 & 1e-4              & 5e-5              & 2.5e-5            \\
Loss weight (CE:MSE) & \multicolumn{4}{c}{0.25:1}                                                       \\
Warmup steps         & 50K                  & 20K               & 1K                & 1K                \\
Training steps       & 200K                 & 200K              & 20K               & 20K               \\
EMA ratio            & -                    & -                 & 0.996             & 0.996             \\
T2I:I2T ratio        & 1.0:0.0              & 0.8:0.2           & 0.8:0.2           & 0.5:0.5           \\
LR schedule          & \multicolumn{4}{c}{Constant}                                                     \\
Weight decay         & \multicolumn{4}{c}{0.0}                                                          \\
Gradient norm clip   & \multicolumn{4}{c}{1.0}                                                          \\
Optimizer            & \multicolumn{4}{c}{AdamW ($\beta_{1}$=0.9,  $\beta_{2}$=0.95, $\epsilon$=1e-15)} \\
\bottomrule
\end{tabular}
\end{table}

%% file: tables/vqa.tex
\begin{table*}[t]
\centering
\caption{Evaluation of multimodal understanding.
Models are categorized based on their architecture types, vision input formats, and input resolution. The reported metrics include accuracy (\%) for Visual Question Answering (VQA), CIDEr scores for image captioning, and scores for general MLLM benchmarks.}
\label{tab:vqa}
\begin{tabular}{lcccc|cc|cc|cccc}
\toprule
                            &           &           &            &      & \multicolumn{2}{c|}{VQA} & \multicolumn{2}{c|}{Captioning} & \multicolumn{4}{c}{General Benchmark} \\
Model                       & \#Params. & Arch.     & Vision     & Res. & VQAv2       & GQA       & Flickr         & COCO          & MME      & POPE   & MMB     & SEED  \\
\midrule
\multicolumn{13}{c}{\textit{Specialist}}                                                                                                                                   \\
\midrule
BEiT-3~\cite{beit3}         & 1.9B      & -         & -          & 224  & 84.0        & -         & -              & 147.6         & -        & -      & -       & -       \\
PaLI-X~\cite{palix}         & 55B       & -         & -          & 256  & 86.1        & -         & -              & 149.2         & -        & -      & -       & -       \\
OFA~\cite{ofa}              & 1B        & -         & -          & 480  & 82.0        & -         & -              & 154.9         & -        & -      & -       & -       \\
\midrule
\multicolumn{13}{c}{\textit{Understanding Generalist}}                                                                                                                     \\
\midrule
BLIP-2~\cite{blip2}                      & 12B       & Q-Former  & Continuous & 224  & 65.0        & 44.7      & 74.9           & \textbf{144.5}         & 1293.8   & 85.3   & -       & 46.4    \\
Flamingo~\cite{flamingo}                    & 80B       & CrossAttn & Continuous & 320  & 56.3        & -         & 67.2           & 84.3          & -        & -      & -       & -       \\
Qwen-VL~\cite{qwen-vl}                     & 9.6B      & Vanilla   & Continuous & 224  & \uline{78.2}        & 57.5      & \uline{81.0}           & -             & 1487.5   & -      & 60.6    & 58.2       \\
LLaVA1.5~\cite{llava1.5}                    & 13B       & Vanilla   & Continuous & 384  & \textbf{80.0}        & \uline{63.3}      & -              & 129.8         & \textbf{1510.7}   & \uline{85.9}   & 64.3    & 58.6    \\
\rowcolor{Tabcolor} \textbf{Libra-1}              & 11B       & Libra     & Hybrid     & 384  & 77.3        & \textbf{63.8}      & \textbf{86.6}           & \uline{135.2}         & \uline{1494.7}   & \textbf{88.2}   & 65.2    & 62.7    \\
\midrule
\multicolumn{13}{c}{\textit{Understanding \& Generation Generalist}}                                                                                                       \\
\midrule
Emu~\cite{emu}              & 14B       & Cascade   & Continuous & 224  & 52.9        & -         & -              & \uline{117.7}         & -        & -      & -       & -       \\
DreamLLM~\cite{dreamllm}    & 7B        & Cascade   & Continuous & 336  & 56.6        & -         & -              & 115.4         & -        &        & 49.9    & -    \\
LaVIT~\cite{lavit}          & 7B        & Cascade   & Discrete   & 224  & 66.0        & 46.8      & \uline{83.0}           & -             & -        & -      & -       & -       \\
NExT-GPT~\cite{nextgpt}     & 13B       & Cascade   & Discrete   & -    & 66.7        & -         & \textbf{84.5}           & \textbf{124.9}         & -        & -      & -       & -       \\
Gemini-Nano-1~\cite{gemini} & 1.8B      & Unified   & -          & -    & 62.7        & -         & -              & -             & -        & -      & -       & -       \\
% VILA-U~\cite{vila-u}        & 7B        & Unified   & Discrete   & 384  & \textbf{79.4}        & 60.8      & -              & -             & 1401.8   & 85.8   & -       & 59.0    \\
Chameleon~\cite{chameleon}  & 34B       & Unified   & Discrete   & -    & 66.0        & -         & 74.7           & -             & -        & -      & -       & -     \\
Janus~\cite{janus}          & 1.3B      & Unified   & Discrete   & 384  & 77.3        & \uline{59.1}      & -              & -             & 1338.0   & 87.0   & \textbf{69.4}    & 63.7    \\
LWM~\cite{lwm}              & 7B        & Unified   & Discrete   & 256  & 55.8        & 44.8      & -              & -             & -        & -      & -       & -     \\
Show-O~\cite{show-o}        & 1.3B      & Unified   & Discrete   & 256  & 59.3        & 48.7      & 36.2           & -             & 948.4    & 73.8   & -       & -       \\
\rowcolor{Tabcolor} & 3B        &  Libra     & Continuous & 256  &     \uline{77.5}        &     58.8      &        78.5        &  115.0             & \uline{1475.0}     &    \uline{88.5}    & 60.0   &    \uline{66.2}
     \\
\rowcolor{Tabcolor} \multirow{-2}{*}{\textbf{Libra-2}}                       & 3B        & Libra     & Continuous & 384  &        \textbf{78.4}     &    \textbf{59.3}      &       77.0         &   107.8            & \textbf{1648.3}   &    \textbf{89.8}    & \uline{60.7}    &     \textbf{71.8}    \\
\bottomrule
\end{tabular}
\end{table*}

%% file: tables/geneval.tex
\begin{table*}
\centering
\caption{Evaluation of text-to-image generation on GenEval benchmark. We present the results of both generation-only methods and methods for unified understanding and generation.}
\label{tab:geneval}
% \resizebox{\textwidth}{!}{
\setlength{\tabcolsep}{3.5mm}{
\begin{tabular}{lcc|ccccccc}
\toprule
Model                          & \#Params & Res. & Single Obj. & Two Obj. & Counting & Colors & Position & Color Attri. & Overal($\uparrow$) \\
\midrule
\multicolumn{10}{c}{\textit{Generation Only}}                                                                                                   \\
\midrule
LlamaGen~\cite{llamagen}       & 0.8B   & 512  & 0.71        & 0.34     & 0.21     & 0.58   & 0.07     & 0.04         & 0.32               \\
LDM~\cite{latent_diffusion}    & 1.4B   &  512 & 0.92        & 0.29     & 0.23     & 0.70   & 0.02     & 0.05         & 0.37               \\
SDv1.5~\cite{latent_diffusion} & 0.9B   &  512 & 0.97        & 0.38     & 0.35     & 0.76   & 0.04     & 0.06         & 0.43               \\
PixArt-$\alpha$~\cite{pixart}  & 0.6B   & 1024  & 0.98        & 0.50     & 0.44     & 0.80   & 0.08     & 0.07         & 0.48               \\
SDv2.1~\cite{latent_diffusion} & 0.9B   & 512  & 0.98        & 0.51     & 0.44     & 0.85   & 0.07     & 0.17         & 0.50               \\
DALL-E 2~\cite{dalle2}         & 6.5B   &  1024 & 0.94        & 0.66     & 0.49     & 0.77   & 0.10     & 0.19         & 0.52               \\
Emu3-Gen~\cite{emu}            & 8B     &  512 & 0.98        & 0.71     & 0.34     & 0.81   & 0.17     & 0.21         & 0.54               \\
SDXL~\cite{sdxl}               & 2.6B   & 1024  & 0.98        & 0.74     & 0.39     & 0.85   & 0.15     & 0.23         & 0.55               \\
\midrule
\multicolumn{10}{c}{\textit{Understanding \& Generation}}                                                                                       \\
\midrule
SEED-X~\cite{seedx}            & 17B    & 256  & 0.97        & 0.58     & 0.26     & 0.80   & \uline{0.19}     & 0.14         & 0.49               \\
Show-o~\cite{show-o}           & 1.3B   & 256  & 0.95        & 0.52     & 0.49     & 0.82   & 0.11     & 0.28         & 0.53               \\
LWM~\cite{lwm}                 & 7B     &  256 & 0.93        & 0.41     & 0.46     & 0.79   & 0.09     & 0.15         & 0.47               \\
Chameleon~\cite{chameleon}     & 34B    &  - & -           & -        & -        & -      & -        & -            & 0.39               \\
Janus~\cite{janus}             & 1.3B   &  384  & 0.97        & 0.68     & 0.30     & 0.84   & \textbf{0.46}     & \textbf{0.42}         & \uline{0.61}               \\
\rowcolor{Tabcolor}    
& 3B     &   256    & \textbf{0.98}        & \textbf{0.71}     & \textbf{0.61}     & \textbf{0.91}   & 0.16     & \uline{0.36}         & \textbf{0.63}               \\
\rowcolor{Tabcolor} \multirow{-2}{*}{\textbf{Libra-2}}                & 3B    & 384   &    \uline{0.97}     &   \uline{0.70}   &  \uline{0.57}    & \uline{0.86}   &  0.11    &    0.28      &     0.58           \\
\bottomrule
\end{tabular}
% }
}
\end{table*}

%% file: tables/fid.tex
\begin{table}
\centering
\caption{Evaluation of text-to-image generation via FID. We present the results of both generation-only methods and methods for unified understanding and generation. $^\dagger$ We build
an image-generation-capable variant of \textbf{Libra-1} by disabling
the contiguous visual signal. }
\label{tab:fid}
% \resizebox{0.5\textwidth}{!}{
% \setlength{\tabcolsep}{1.0mm}{
\begin{tabular}{lccc}
\toprule
Model                             & \#Params & COCO-30K($\downarrow$) & MJHQ-30K($\downarrow$) \\
\midrule
\midrule
\multicolumn{4}{c}{\textit{Generation Only}}                                                           \\
\midrule
DALL-E~\cite{dalle}               & 12B     & 27.50                  & -                      \\
GLIDE~\cite{glide}                & 5B      & 12.24                  & -                      \\
LDM~\cite{latent_diffusion}       & 1.4B    & 12.64                  & -                      \\
DALL-E 2~\cite{dalle2}            & 6.5B    & 10.39                  & -                      \\
SDv1.5~\cite{latent_diffusion}    & 0.9B    & 9.62                   & -                      \\
GigaGAN~\cite{gigagan}            & 0.9B    & 9.09                   & -                      \\
PixArt-$\alpha$~\cite{pixart}     & 0.6B    & 7.32                   & -                      \\
Imagen~\cite{imagen}              & 34B     & 7.27                   & -                      \\
RAPHAEL~\cite{raphael}            & 3B      & 6.61                   & -                      \\
\midrule
\multicolumn{4}{c}{\textit{Understanding \& Generation}}                                                \\
\midrule
Emu~\cite{emu}          & 13B     & 11.66                  & -                      \\
NExT-GPT~\cite{nextgpt} & 13B     & 11.28                  & -                      \\
SEED-X~\cite{seedx}     & 17B     & 14.99                  & -                      \\
Show-o~\cite{show-o}              & 1.3B    & 9.24                   & 15.18                  \\
LWM~\cite{lwm}                    & 7B      & 12.68                  & 17.77                  \\
VILA-U~\cite{vila-u}              & 7B      & -                      & 12.81                  \\
Janus~\cite{janus}                & 1.3B    & 8.53                   & \textbf{10.10}                  \\
\rowcolor{Tabcolor} \textbf{Libra-1}$^{\dagger}$                   & 11B      & 49.74                   & 71.06                  \\
\rowcolor{Tabcolor} \textbf{Libra-2} ($256^{2}$)                   & 3B      & \uline{8.47}                   &  11.58                  \\
\rowcolor{Tabcolor} \textbf{Libra-2} ($384^{2}$)                   & 3B      & \textbf{8.13}                   &  \uline{10.39}                  \\
\bottomrule
\end{tabular}
% }
% }
\end{table}

%% file: tables/ablation_libra.tex
\begin{table}[t]
\centering
\caption{Ablation of the basic architecture design on \textbf{Libra-1}.}
\label{tab:ablation_libra}
\resizebox{\linewidth}{!}{
\setlength{\tabcolsep}{0.8mm}{
\begin{tabular}{ccclc|c|ccc}
\toprule
Exp          & Ablated  & Original   & \multirow{2}{*}{$\rightarrow$} & Changed      & \multirow{2}{*}{\#Params} & \multicolumn{3}{c}{MLLM Benchmark} \\
ID           & Setting  & Value      &                                & Value        &                           & MME         & POPE      & SEED     \\ \midrule 
\rowcolor{Tabcolor}             & \multicolumn{4}{c|}{\textbf{Original Libra-1 Model}}                             & 11.3B                     & 1494.7      & 88.2      & 62.7     \\ \midrule
\textbf{(a)} & Architecture & Libra      &                                & Unified      & 7.5B                      & 1450.2      & 85.9      & 58.4     \\
\textbf{(b)} & Bridge   & \Checkmark &                                & \XSolidBrush & 11.2B                     & 1458.4      & 86.0      & 59.6     \\
\textbf{(c)} & Tokenization    & Hybrid     &                                & Discrete     & 11.3B                     & 1127.8      & 80.3      & 48.5     \\
% \textbf{(d)} & Encoder  & CLIP       &                                & Scratch      & 11.3B                     & 1148.4      & 80.7      & 50.8     \\ 
\textbf{(d)} & Supervision  & Unified       &                                & Language      & 11.3B                     & 1465.5      & 86.0      & 58.8     \\ 
\bottomrule
\end{tabular}
}
}
\end{table}

%% file: tables/ablation.tex
\begin{table}[t]
\centering
\caption{Ablation of the improvements for unified multimodal understanding and generation on \textbf{Libra-2}.}
\label{tab:ablation}
\setlength{\tabcolsep}{1.7mm}{ 
\begin{tabular}{ccclc|cc}
\toprule
Exp ID & Details      & Original   & $\rightarrow$ & Target       & MME & GenEval \\
\midrule
\rowcolor{Tabcolor}       & \multicolumn{4}{c|}{\textbf{Original Libra-2 Model}}       &  1475   &     0.63    \\
\midrule
\textbf{(a)}      & Switch FFN   & \Checkmark &               & \XSolidBrush &   1406  &  0.61       \\
\textbf{(b)}      & UniRoPE      & \Checkmark &               & \XSolidBrush &   1252  &  0.42       \\
\textbf{(c)}      & Tokenization & Continuous &               & Hybrid     &   1416  &  -       \\
\bottomrule
\end{tabular}
}
\end{table}

%% file: tables/ablation_obj.tex
\begin{table}[t]
\centering
\caption{Analysis on the mutual impact of understanding and generation within \textbf{Libra2}.}
\label{tab:gen_vs_und}
\setlength{\tabcolsep}{1.8mm}{
\begin{tabular}{l|ccc|ccc}
\toprule
\multirow{2}{*}{Method} & \multicolumn{3}{c|}{Understanding} & \multicolumn{3}{c}{Generation} \\
                        & VQAv2       & MME      & MMB      & COCO    & MJHQ    & GenEval    \\
\midrule
Und. Only               &      76.0       &   \textbf{1513}    &    58.2  &   -      &     -    &     -       \\
Gen. Only               &      -       &     -     &     -     &     8.79    &   12.14      &      0.53      \\
\rowcolor{Tabcolor} Und. \& Gen.            &   \textbf{77.5}          &    1475      &      \textbf{60.0}    &     \textbf{8.47}    &    \textbf{11.58}     &      \textbf{0.63}      \\
\bottomrule
\end{tabular}
}
\end{table}

%% file: sections/conclusion.tex
\section{Conclusion}
% \textbf{Conclusion}.
Through developing the Libra series, including \textbf{Libra-1} and \textbf{Libra-2}, we show that a decoupled vision-language system can improve both multimodal understanding and generation. The Libra architecture achieves this via switch attention and switch FFN modules. We observe that this design yields diverse attention patterns across layers, suggesting reduced learning redundancy. In addition, we integrate continuous-space masked image generation into a unified MLLM, which endows \textbf{Libra-2} with both accurate multimodal understanding and high-quality image generation. We would like to claim that vision and language should be integrated in a more reasonable manner beyond simple modality alignment. We hope our work could provoke more consideration in MLLM designs.

% \textbf{Limitations}.
% lack foundation support
% 